\pdfoutput=1

\documentclass[11pt]{article}

\PassOptionsToPackage{table}{xcolor}
\usepackage{acl}

\usepackage{times}
\usepackage{latexsym}
\usepackage{adjustbox}

\usepackage[T1]{fontenc}

\usepackage[utf8]{inputenc}
\usepackage{enumitem}

\usepackage{microtype}

\usepackage{inconsolata}

\usepackage{multirow} 
\usepackage{graphicx}
\usepackage{booktabs}
\usepackage{float}
\usepackage{makecell} 

\usepackage{amsmath}
\usepackage{amssymb}

\usepackage[most]{tcolorbox}
\usepackage{geometry}
\usepackage{caption}
\usepackage{cuted}     
\usepackage{multirow}  
\usepackage{booktabs}  
\usepackage{tcolorbox}
\usepackage{textcomp}

\definecolor{appmeans}{HTML}{D94F4F}     
\definecolor{modelmeans}{HTML}{3A8C8C}   
\definecolor{ends}{HTML}{6C8EBF}         
\definecolor{actor}{HTML}{7D60A6}        
\definecolor{subject}{HTML}{555555}  
\definecolor{ClearBlue}{RGB}{0, 60, 140}
\definecolor{VagueOrange}{RGB}{255, 130, 130} 

\usepackage{tablefootnote}
\usepackage{longtable}
\usepackage{cleveref}
\usepackage{multirow}
\usepackage{xspace}
\usepackage{linguex}

\usepackage{tabularx}

\usepackage{inconsolata}

\usepackage{makecell}
\definecolor{lightred}{RGB}{255,228,225}
\definecolor{lightblue}{RGB}{194, 200, 255}

\title{Social Good or Scientific Curiosity? \\ Uncovering the Research Framing Behind NLP Artefacts}

\author{
Eric Chamoun\textsuperscript{1}, 
Nedjma Ousidhoum\textsuperscript{2}\thanks{Equal contribution.}, 
Michael Schlichtkrull\textsuperscript{3,*}, 
Andreas Vlachos\textsuperscript{1} \\
\textsuperscript{1}Department of Computer Science and Technology, University of Cambridge \\
\textsuperscript{2}Cardiff University \\
\textsuperscript{3}Queen Mary University of London \\
\texttt{\{ec806,av308\}@cam.ac.uk}, \texttt{ousidhoumN@cardiff.ac.uk}, \texttt{m.schlichtkrull@qmul.ac.uk}
}

\begin{document}
\maketitle
\begin{abstract}
Clarifying the research framing of NLP artefacts (e.g., models, datasets, etc.) is crucial to aligning research with practical applications when researchers claim that their findings have real-world impact. 
Recent studies manually analyzed NLP research across domains, showing that few papers explicitly identify key stakeholders, intended uses, or appropriate contexts. 
In this work, we propose to automate this analysis, developing a three-component system that infers research framings by first extracting key elements (means, ends, stakeholders), then linking them through interpretable rules and contextual reasoning.
We evaluate our approach on two domains: automated fact-checking using an existing dataset, and hate speech detection for which we annotate a new dataset\footnote{Code and annotations available at our \href{https://github.com/ericchamoun/NLP-Framing-Analysis}{GitHub repository}.}
—achieving consistent improvements over strong LLM baselines.
Finally, we apply our system to recent automated fact-checking papers and uncover three notable trends: a rise in underspecified research goals, increased emphasis on scientific exploration over application, and a shift toward supporting human fact-checkers rather than pursuing full automation.

\end{abstract}
\section{Introduction}
As NLP systems see wider real-world deployment, concerns over misuse and unintended consequences have intensified focus on responsible AI practices—highlighting the need for ethical integration, risk mitigation, and attention to societal impact \citep{noble, bender, birhane}.
A critical but often overlooked aspect of responsible NLP research is \textit{research framing}—how a paper positions the purpose and value of its work by connecting \textit{what} it does (means), \textit{why} it does it (ends), and for \textit{whom} it is relevant (stakeholders). 
In this context, framing highlights the methodological specification of an artefact’s role within the research ecosystem. This contrasts with political or media framing, which emphasizes persuasion in discourse, and semantic framing, which concerns meaning in linguistic structure \cite{card-etal-2016-analyzing}. 

When poorly articulated, research framings can obscure a system’s scope, hinder accountability, and increase the risk of misuse.
%
Figure~\ref{fig:framing_examples} illustrates this point by comparing two NLP papers: one clearly intends to support professional fact-checkers with specified means, ends, and users; the other, on hate speech detection, lacks any mention of who would use the model’s outputs. Such omissions risk misinterpretation—e.g., framing a label-only classification model as an automated decision tool overlooks the need for context and explanation, risking misuse in high-stakes settings.

\begin{figure*}[t]
\centering
\footnotesize

\begin{adjustbox}{minipage=[t]{0.48\textwidth},valign=t}
\begin{tcolorbox}[colback=white, colframe=black, width=\linewidth, boxsep=2pt, left=2pt, right=2pt, title=Assistive deployment (Fact-checking)]
\textbf{Excerpt:} In an effort \textcolor{actor}{\textbf{to assist fact-checkers}}, we tackle the \textcolor{appmeans}{\textbf{claim detection task}}... \textcolor{ends}{\textbf{Misinformation}} has recently become \textcolor{ends}{\textbf{more central in public discourse}} ... NLP approaches to \textcolor{ends}{\textbf{alleviate}} the activity of \textcolor{ends}{\textbf{fact-checking}}. We introduce an approach ... \textcolor{modelmeans}{\textbf{to perform the classification.}} 
\vspace{0.2cm}\\
\textbf{Epistemic elements}
\begin{itemize}[noitemsep, topsep=1pt, leftmargin=*]
    \item \textbf{Model means:} \textcolor{modelmeans}{What (ML) methods are proposed?} → classify/score veracity
    \item \textbf{Application means:} \textcolor{appmeans}{How will the artefact be used?} → identify claims
    \item \textbf{Data actors:} \textcolor{actor}{Who will act on the predictions from the artefact?} → professional fact-checkers
    \item \textbf{Ends:} \textcolor{ends}{What is the intended societal goal?} → fight misinformation
\end{itemize}
\vspace{0.2cm}
\textbf{Framing:} Clear assistive framing, artefacts support journalists.
→ 
\textbf{Assisted external fact-checking}.
\end{tcolorbox}
\end{adjustbox}
\hfill
\begin{adjustbox}{minipage=[t]{0.48\textwidth},valign=t}
\begin{tcolorbox}[colback=white, colframe=black, width=\linewidth, boxsep=2pt, left=2pt, right=2pt, title=Vague deployment (Hate speech)]
\textbf{Excerpt:} Current filters are \textcolor{ends}{\textbf{insufficient to prevent the spread of hate speech.}} Most \textcolor{subject}{\textbf{internet users}}... report having been subjected to offensive name-calling... leverage our annotations to improve \textcolor{appmeans}{\textbf{hate speech detection}}... \textcolor{modelmeans}{\textbf{Treating hate speech classification}} ...  \textcolor{modelmeans}{\textbf{label}}
around 13,000 potentially \textcolor{modelmeans}{\textbf{derogatory tweets}}.
\vspace{0.2cm}\\
\textbf{Epistemic elements}
\begin{itemize}[noitemsep, topsep=1pt, leftmargin=*]
    \item \textbf{Model means:} \textcolor{modelmeans}{What (ML) methods are proposed?} → classify/score posts/text, data collection
    \item \textbf{Application means:} \textcolor{appmeans}{How will the artefact be used?} → identify toxic content
    \item \textbf{Data actors:} \textcolor{actor}{Who will act on the predictions from the artefact?} → not specified
    \item \textbf{Ends:} \textcolor{ends}{What is the intended societal goal?} → fight hate
\end{itemize}
\vspace{0.2cm}
\textbf{Framing:} Clear goals and methods, but the user of the model’s outputs is unspecified. → \textbf{Vague identification}.
\end{tcolorbox}
\end{adjustbox}

\caption{Side-by-side examples of research framings. The fact-checking paper (left, \citet{fc-example-new}) presents a clear assisted external fact-checking framing with specified users, goals, and methods. The hate speech paper (right, \citet{hatespeech-example}) has an underspecified framing due to unspecified model users. }
\label{fig:framing_examples}
\end{figure*}
This concern has motivated several analyses of how NLP research is framed. \citet{intendeduses} first formalized the notion of research framings (which they termed ``epistemic narratives'') and introduced \textit{epistemic elements}—textual signals that define a paper’s framing by specifying components such as means, ends and stakeholders. Analyzing 100 automated fact-checking papers, they found frequent misalignments between goals and methods, underspecified or missing stakeholder definitions, and weak links between these elements. Separately, \citet{liu-etal-2023-responsible} reviewed 333 summarization papers and found that fewer than 15\% meaningfully addressed responsible AI concerns. 
 
Such analyses help track field-wide trends to inform future research and reveal problems in research practices, including value misalignments and overlooked downstream risks \citep{birhane}.
They further offer guidance by encouraging alignment between research goals and stakeholder needs, prompting reflection on issues like dual use \citep{leins2020} and overclaiming \citep{grodzinsky}. However, their reliance on manual annotation limits their scalability and timeliness.

In this work, we first propose a generalizable research framing schema to support cross-domain analysis in NLP (\S\ref{sec:cross-domain}). Building on this, we introduce an automated framework for inferring research framings (\S\ref{sec:approach}), composed of three stages:
\begin{enumerate}[noitemsep, topsep=0pt, leftmargin=*]
    \item \textit{Epistemic element extraction}, i.e., identifying parts of the introductory text that specify the components of a paper’s research framing, such as its aims, means and stakeholders.
    \item \textit{Research framing ranking}, which uses semi-automatically inferred rules to connect extracted elements, linking means to ends to infer the artefact’s intended purpose;
    \item \textit{Research framing classification}, which leverages an LLM to refine rankings with broader context, in-context examples, and reasoning.
\end{enumerate}


We evaluate our approach (\S\ref{sec:results}) in two domains: automated fact-checking (AFC) and hate speech detection (HS). For AFC, we use the dataset released by \citet{intendeduses}; for HS, we manually annotate 49 highly cited papers (\S\ref{sec:annotation}) with epistemic elements, which we use to infer their research framings.
Our approach outperforms a strong LLM baseline in both domains, offering more accurate research framing predictions. It also achieves high precision in identifying underspecified framings with misaligned or incomplete elements. 


Finally, we apply our system (\S\ref{sec:meta-analysis}) to recent AFC papers to uncover emerging trends. 
We observe a rise in underspecified research framings, raising concerns about the clarity of current framing practices in AFC research. We also find an increase in papers categorized under \textit{Scientific Curiosity}, largely driven by benchmark studies examining LLM limitations. Lastly, we find a growing emphasis on generating fact-checking justifications to assist human fact-checkers, while fewer studies propose fully automating the process.

\textbf{Research framings for this paper}
We release two NLP artefacts: (1) datasets used for analysis and (2) a tool for generating research framing labels and justifications (\textit{modeling means}). Both are designed for use by researchers (\textit{data actors}, \textit{model owners}), who are also the \textit{data subjects}. The tool is intended to support large-scale analyses of NLP subfields and provide feedback on how artefacts are positioned for use or exploration (\textit{application means}). The study aims to advance NLP knowledge and promote framing transparency (\textit{ends}), aligning with the research framings of \textit{Scientific Curiosity} and \textit{Scientific Writing Assistance}. 

\renewcommand{\arraystretch}{0.9}
\begin{table*}[t]
\centering
\small
\begin{tabular}{p{3.1cm}p{6.1cm}p{2.5cm}p{2.5cm}}
\toprule
\textbf{General Framing} & \textbf{Description} & \textbf{AFC} & \textbf{HS}  \\
\midrule
\textcolor{ClearBlue}{\textbf{Automated deployment}} & System replaces a human task with minimal intervention. & Automated external fact-checking & Automated content moderation \\
\textcolor{ClearBlue}{\textbf{Assistive deployment}} & System supports human decision-making. & Assisted internal/external fact-checking & Assisted content moderation \\
\textcolor{ClearBlue}{\textbf{Knowledge access and curation}} & Organizes/synthesizes knowledge for future use. & Assisted knowledge curation & Assisted knowledge curation  \\
\textcolor{ClearBlue}{\textbf{Knowledge exploration}} & Explores models or data without specific application goals. & Scientific curiosity & Scientific curiosity \\
\textcolor{ClearBlue}{\textbf{Governance}} & Supports legal, institutional, or compliance goals. & Law enforcement & Law enforcement \\
\midrule
\textcolor{VagueOrange}{\textbf{Vague deployment}} & Implies deployment but omits how or where the model is used. & Vague debunking & Vague moderation  \\
\textcolor{VagueOrange}{\textbf{Vague opposition}} &States a broad goal, but lacks a coherent link between that goal and the proposed ML method.  & Vague opposition & Vague opposition \\
\makecell[l]{\textbf{\textcolor{VagueOrange}{Vague identification}} \\ \textcolor{VagueOrange}{\textbf{(Detection tasks)}} } & Identifies content without specifying how it is used to achieve stated ends and who acts on it. & Vague identification & Vague identification  \\

\bottomrule
\end{tabular}

\caption{Cross-domain research framings, with corresponding instantiations in AFC, HS. Clear framings (blue) link technical contributions to explicit purposes/users; underspecified framings (pink) lack specificity about use/users/goals.}
\label{tab:research-framings}
\end{table*}

\section{Cross-domain research framing}
\label{sec:cross-domain}
A research framing captures how a paper connects key epistemic elements to convey the intended purpose of its contributions. We consider the following epistemic elements: 
(1) \textit{data subjects}—individuals whose behaviors are analyzed (e.g., social media users), (2) \textit{data actors}—those intended to use the model outputs (e.g., moderators), (3) \textit{model owners}—entities controlling deployment (e.g., social media companies), (4) \textit{modeling means}—the machine learning approach (e.g., classification), (5) \textit{application means}—how the model is used (e.g., automated removal), and (6) \textit{ends}—the intended societal or research goal (e.g., fighting hate online).
These elements (listed in Appendix~\ref{appendix:universe}) may be explicitly connected, partially specified, or mismatched—e.g., presenting a classifier as a replacement for human experts without accounting for justification or user context.

To support automated analysis across domains, we propose a generalizable research framing schema. Table~\ref{tab:research-framings} presents our cross-domain framing types, along with their instantiations in AFC \citep{intendeduses} and HS (this work). While our evaluation focuses on these two domains, the schema is designed to be extensible;  e.g., mental health applications like \textit{automated monitoring} or \textit{mandatory flagging} can align  with this structure.

To assess framing quality, we distinguish between \textit{clear} and underspecified cases. Clear framings specify the artefact’s purpose, users, and goals; underspecified ones omit key components or introduce inconsistencies that obscure the system’s intended use. 
We define underspecification relative to how clearly authors articulate and connect epistemic elements at the time of writing. In other words, it reflects presentation clarity in its contemporary context, not compliance with future or evolving community standards. This time-bounded definition makes the schema generalizable, since it provides a consistent basis for analyzing papers across different publication periods.

We also note that theoretical research without an immediate ends or intended user is not the same as underspecification. If a paper clearly states that the goal is theoretical exploration or to answer some research question, that corresponds to \textbf{Knowledge exploration}, a clear framing. Underspecification arises when this intent is unstated, e.g., authors claiming that their theoretical findings aim to \textit{``fight hate''} without discussing how.

\noindent We highlight hate speech-specific framings (see Appendix~\ref{appendix:universe-fc} for those specific to AFC):

\noindent \textbf{LLM safety }represents a clear framing focused on preventing harmful LLM outputs through detoxification, filtering, or alignment. 
\textbf{Vague data analysis} refers to papers combining knowledge exploration with social aims (e.g.,``fighting hate''), without a clear link between analysis and real-world impact.

\section{Inference of Research Framings}
\label{sec:approach}

We automatically infer the \textit{research framings} in NLP papers 
in three stages
(Figure \ref{fig:task-overview}):

\begin{figure*}
        \centering
        \includegraphics[width=\linewidth]{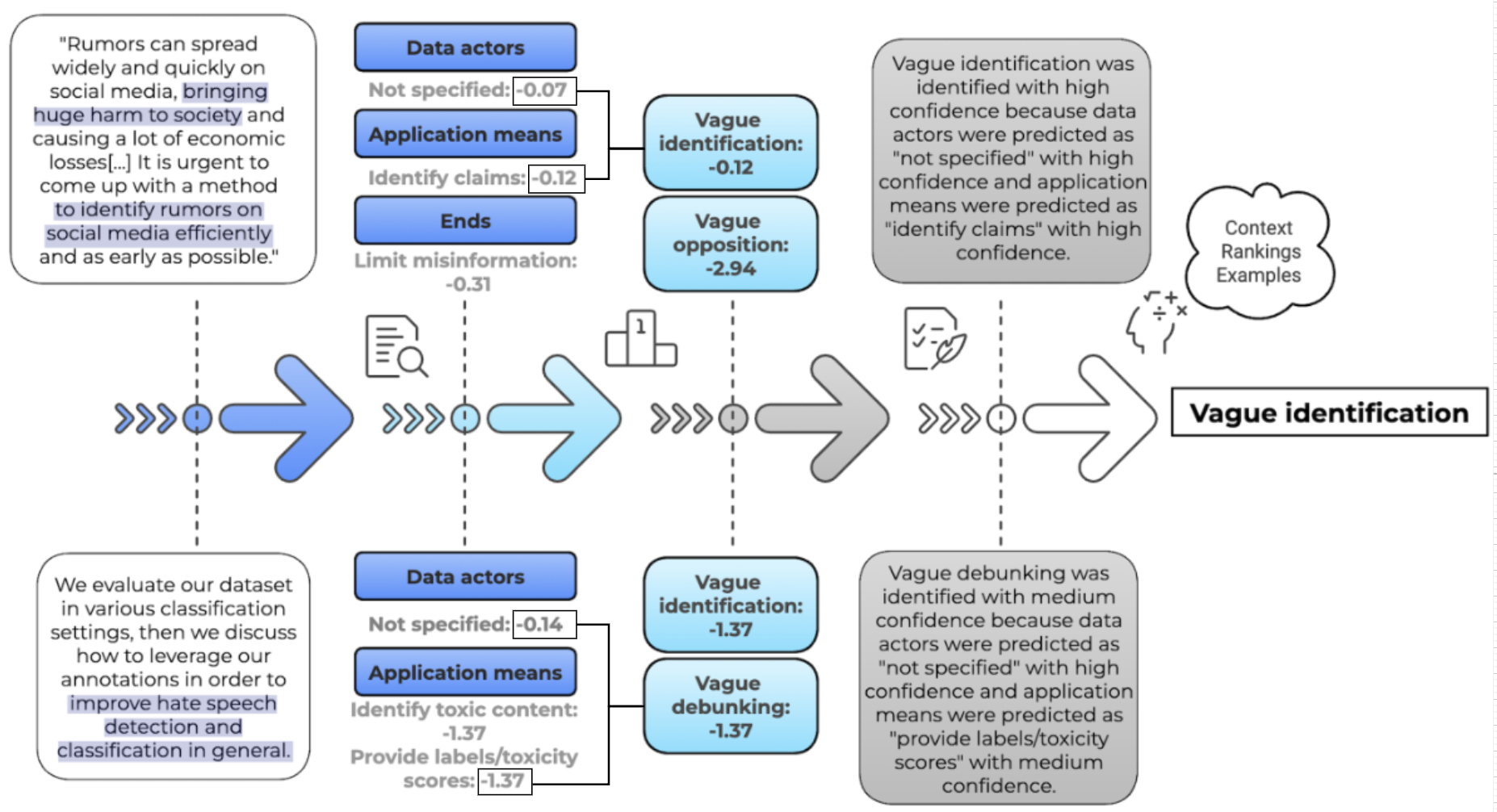}
        \caption{Our system infers a paper's research framings by: (1)\ estimating epistemic element likelihoods; (2)\ applying logical rules linking elements to framings; (3)\ generating explanations summarizing the likelihoods computed in steps 1 and 2  as reasoning for classification; (4)\ using an LLM to refine predictions using the explanations, context, and in-context examples. Numbers represent log-likelihoods of predicted elements or framings. Examples are from \citet{fc-example} (top) and \citet{hatespeech-example} (bottom). }
        \label{fig:task-overview}
    \end{figure*}
\subsection{Step 1: Epistemic Element Extraction}
\label{sec:element-extraction}


We treat element extraction as a multi-label classification task, since a paragraph may contain multiple relevant elements (e.g., several stakeholders or goals). We also allow the model to abstain when no evidence is present for any element category.

\paragraph{Formulation.} 
Given a paragraph $p_i$ from a selected paper and an epistemic element  $E_j \in \mathcal{E}$ (e.g. \textit{data actors}), we define the corresponding candidate label set $\mathcal{R}(E_j)$  (including \textit{not specified}). We frame the extraction as a QA task, using a language model $M$ to compute response likelihoods:\\
\begin{equation}
s_{p_i, r_{E_j,k}} = \log P_M(r_{E_j,k} \mid p_i, q_{E_j})
\end{equation}
where $q_{E_j}$ is a predefined prompt targeting element $E_j$ (e.g., \textit{``Who are the individuals or entities that the authors identify as expected to act on the released artefacts?''}) and $r_{E_j,k} \in \mathcal{R}(E_j)$ is a candidate label (e.g. \textit{journalists}).

\paragraph{Semantic Clustering via NLI.}
We adapt \citet{semantic-entropy} to cluster sampled outputs $\{x_1, \ldots, x_n\}$ into semantically equivalent classes $C = \{c_1, \ldots, c_m\}$. Two outputs $x_a, x_b$ are grouped together if they entail each other bidirectionally (i.e., NLI returns \texttt{entail} in both directions). Class likelihoods are computed via:
\begin{equation}
s_{c_j} = \log \sum_{x \in c_j} \exp(s_{p_i, x})
\end{equation}
\subsection{Step 2: Research Framing Ranking}
\label{sec:framing-ranking}
In this step, we estimate the likelihood of each research framing $f_i \in \mathcal{F}$ by linking predicted epistemic elements through interpretable logical rules.
Each research framing \( f_i \) is defined by a boolean rule $\mathcal{L}_i : \{r_{E_1,k_1}, r_{E_2,k_2}, \dots, r_{E_m,k_m}\} \implies f_i$ that specifies how combinations of epistemic elements must co-occur to support that framing. These rules are constructed from two sources: (1) a CART-based decision tree \cite{cart} trained on annotated data, and (2) manual refinement based on framing definitions (Appendix~\ref{logical-mappings-appendix}). 
An example is shown in Figure~\ref{fig:tree-to-rule} in the appendix.

To compute the likelihood of a research framing, we evaluate each logical rule \( \mathcal{L}_i \) over the predicted probabilities of its constituent epistemic elements. Conjunctions and disjunctions are computed as \[
\Pr(r_{E_a,k} \land r_{E_b,l}) = \min\{\Pr(r_{E_a,k}), \Pr(r_{E_b,l})\},
\]
\[
\Pr(r_{E_a,k} \lor r_{E_b,l}) = \max\{\Pr(r_{E_a,k}), \Pr(r_{E_b,l})\}.
\].
This soft logic enables smooth aggregation of partial evidence across elements.

At inference time, we aggregate paragraph-level predictions to compute paper-level scores for each epistemic element. Since research framings describe the broader intended purpose of an NLP artefact, these elements must be assessed across the full document rather than in isolated paragraphs.

Let a paper consist of paragraphs $\{p_1, p_2, \dots, p_T\}$, and let 
$\Pr(r_{E_j,k} \mid p_t)$ denote the confidence score of label $r_{E_j,k}$ for element 
$E_j$ in paragraph $p_t$. We compute the paper-level score as:
\[
\Pr(r_{E_j,k}) = \max_{t=1}^{T} \Pr(r_{E_j,k} \mid p_t).
\]

To model absence, we define the paper-level score for the \textit{not specified} label 
of element $E_j$ as:
\[
\Pr(r_{E_j,\text{ns}}) = \min_{t=1}^{T} \Pr(r_{E_j,\text{ns}} \mid p_t).
\]
This aggregation ensures that element presence is acknowledged if mentioned anywhere, while absence requires consistent omission across
text.
\subsection{Research Framing Classification}

To refine framing predictions, we use an LLM $M$ to incorporate document context and reason beyond deterministic rules.
We first collect confidence scores from previous stages. Each framing $f_i$ is 
associated with a confidence score $\Pr(f_i)$, and its supporting epistemic 
element labels $S_i$, $\quad \text{as specified by } \mathcal{L}_i$, have scores 
$\Pr(r_{E_j,k})$. We discretize each score into a confidence tier for each paper:
\begin{align}
\text{Tier}(\Pr(z)) &=
\begin{cases}
\text{High}   & \text{if } \Pr(z) \geq \tau_h \\
\text{Medium} & \text{if } \tau_m \leq \Pr(z) < \tau_h \\
\text{Low}    & \text{if } \tau_l \leq \Pr(z) < \tau_m
\end{cases} \notag \\
&\quad \text{for all } z \in \{f_i\} \cup S_i \notag
\end{align}
where \( \tau_h \), \( \tau_m \), \( \tau_l \) are predefined confidence thresholds. We then generate structured natural language justifications summarizing the prediction rationale: \textit{``[Framing] was identified with [confidence level] because [element 1] was predicted with [confidence level X], [element 2] with [confidence level Y], ...''}

$M$ is fed explanations along with paper context, framing definitions, and examples (Appendix~\ref{prompts-appendix}).

\section{Research Framings in Hate Speech}
\label{sec:annotation}
To assess cross-domain generalization, we apply our approach to hate speech detection—a task often framed as NLP for Social Good \cite{nlp4sg}, where artefacts aim to assist moderators in combating online harm.
We construct a new dataset of 49 highly cited papers on HS and related tasks from top NLP and CSS venues, annotated for epistemic elements and discourse-level research framings. 

\paragraph{Annotation Procedure.} We follow a two-phase protocol based on \citet{krippendorff}: 
\begin{enumerate}[noitemsep, topsep=0pt, leftmargin=*]
\item Initial annotation used a predefined taxonomy of epistemic elements and research framings inspired by prior work \cite{intendeduses}, with new categories added inductively as needed. Annotators met regularly to discuss and reconcile divergent cases. 
\item A second round re-annotated all papers using the unified, finalized label set. No new labels were added in this phase, ensuring convergence on a minimal, interpretable set of research framing types.
\end{enumerate}
Annotations were performed by two NLP researchers with domain expertise in hate speech and sociotechnical NLP. Inter-annotator agreement for the final label set reached a Krippendorff's $\alpha$ of 0.61 across the research framing labels (10 distinct research framings with multiple per paper).

\paragraph{Annotation Scope.} For each paper, we extract quotes from the abstract and introduction that describe the system’s purpose, methods and stakeholders. We annotate the presence of epistemic element types defined in \S \ref{sec:cross-domain}. Each paper is then assigned one or more discourse-level \textit{research framings}. Appendix~\ref{new-narratives-appendix} defines the framing types.
%
All examples were annotated based on explicitly stated text only, avoiding inferred intent. Detailed guidelines are provided in Appendix \ref{annotation-appendix}.

\paragraph{Findings}
The distribution of research framings in our annotated HS dataset reveals that \textit{Vague Identification} is the most prevalent (44.74\%). These papers present systems for detecting toxic content but fail to specify who will act on the outputs or how. Overall, underspecified research framings make up 64.47\% of the dataset, underscoring a broader trend of ambiguity in the intended use of hate speech detection models (see Appendix~\ref{app:hs-dist} for distribution details). This is reinforced by the prevalence of \textit{Vague Data Analysis}, where papers combine knowledge exploration with social aims (e.g., ``fighting hate'') without articulating how the analysis supports real-world impact.

%
The second most common framing is \textit{Scientific Curiosity}, often seen in papers that examine the linguistic characteristics of hate speech—such as nuance and sarcasm—or that introduce new datasets and benchmarks. Notably, only around 15\% of papers clearly specify practical applications for mitigating hate speech, with a small percentage (2.63\%) focusing specifically on limiting toxic content generated by LLMs.




\section{Experimental setup}
\label{sec:results}

\subsection{Implementation details} 
\paragraph{Modeling} We use \texttt{gemini-1.5-flash-002} \citep{gemini} for all prompting tasks, including epistemic element extraction (Step 1) and research framing classification (Step 3). Gemini has demonstrated strong performance on open-ended tasks like writing assistance, balancing diversity and contextual alignment \cite{writingastestbed}.
We provide all full prompts in Appendix~\ref{prompts-appendix}.

\paragraph{Epistemic Element Extraction}
For each paragraph, we generate \( k = 10 \) samples per epistemic element, following \citet{semantic-entropy}, as increasing samples improves performance (Appendix~\ref{semantic-clustering-val}). 
We set the temperature to 1.0, balancing diversity and coherence—lower values produce overly deterministic responses, while higher ones risk incoherence.
%
Clustering is performed using NLI with DeBERTa  \cite{deberta} to merge semantically equivalent responses. Entailment is determined by the model’s predicted label.

\paragraph{Framing Ranking} To map epistemic elements to research framings, we employ two decision trees: (1) a global tree that identifies key elements distinguishing research framings and (2) individual trees per research framing, capturing characteristic element combinations in a multi-label setting. Nodes are split using Gini impurity, with the global tree fully expanded for differentiation and individual trees capped at depth 4 for interpretability. Trees are trained using Gini impurity as the splitting criterion. The resulting paths are converted into boolean rules (see Section~\ref{sec:framing-ranking} and Appendix~\ref{logical-mappings-appendix}).

\paragraph{Absence Thresholding} Papers often present multiple research framings across sections—e.g., one section may frame a tool as identifying misleading claims without specifying intended actors (\textit{Vague Identification}), while another specifies its use by \textit{journalists}. Strictly assigning -$\infty$ to \textit{not specified} labels when \textit{data actors} appear in some paragraphs would incorrectly exclude \textit{Vague Identification}, even if applicable elsewhere. To balance specificity with flexibility, we set a -3 threshold for \textit{not specified} labels—low enough to avoid inflating irrelevant elements (as model-assigned likelihoods exceed -3) but high enough to retain research framings dependent on their absence. 

\paragraph{Framing Classification}

We define confidence tiers for both epistemic element and framing scores using the thresholds \( \tau_h = -1 \), \( \tau_m = -2 \), and \( \tau_l = -3 \), consistent with Gemini’s typical scoring range of approximately \([-3, 0]\). These tiers are used to guide natural language justifications, which are passed to the LLM in conjunction with paper context, framing definitions, and in-context examples (see Figure~\ref{fig:task-overview}). Using natural language labels for confidence levels improves the model’s reasoning by making the signal more interpretable. 


\subsection{Evaluation metrics}
We evaluate performance for \textbf{epistemic element extraction} and \textbf{research framing ranking} using Filtered Mean Reciprocal Rank (fMRR; \citet{fmrr}), which averages the inverse rank of correct predictions while excluding higher-ranked known positives. That is, when multiple correct answers exist, we compute the rank of a target answer after removing other correct answers ranked above it.
%
Also, some labels are
unscored ($-\infty$) if not predicted (e.g., \textit{professional journalists} when only \textit{social media users} are mentioned in the paper). 

For scored labels, we compute standard fMRR. If a correct label is unscored, we estimate its rank as the number of scored labels plus half the unscored ones, assuming unscored items are randomly distributed. For example, with 2 scored and 10 unscored labels, the estimated rank is $2 + \frac{10}{2} = 7$, preventing artificially inflated results.
\begin{align}
\text{fMRR} &= \frac{1}{N} \sum_{i=1}^{N} \frac{1}{\text{rank}_i} \\
\text{rank}_i &=
\begin{cases}
r_i, & \text{if label } i \text{ is scored} \\
s + \frac{u}{2}, & \text{if label } i \text{ is unscored}
\end{cases}
\end{align}
\noindent where \( N \) is the number of evaluated instances, \( r_i \) the observed rank of the correct label, \( s \) the number of scored labels, \( u \) the number of unscored labels.

\noindent For \textbf{research framing classification}, we report the Jaccard score, i.e., the intersection over union of predicted and gold labels:
\(
\text{Jaccard}(A, B) = \frac{|A \cap B|}{|A \cup B|}
\) \cite{jaccard}.
It is well-suited to our multi-label setting, where few labels are typically present among 10+ candidates. Unlike F-score—which can be inflated by correctly predicting absent labels—Jaccard evaluates only active labels.
We also report F-score for specific framings for completeness.

\subsection{Epistemic element extraction}

\begin{table}[ht]
\centering
\resizebox{\linewidth}{!}{
\begin{tabular}{@{}lcc|cc||cc@{}}
\toprule
\textbf{} & \multicolumn{4}{c}{\textbf{AFC}} & \multicolumn{2}{c}{\textbf{HS}} \\

\textbf{} & \multicolumn{2}{c}{\textbf{Validation}} & \multicolumn{2}{c}{\textbf{Test}} &  \\ 
 & \textbf{All} & \textbf{Oracle} & \textbf{All} & \textbf{Oracle} & \textbf{All} & \textbf{Oracle} \\ 
\midrule
Data Subjects     & \textbf{73.80} & 72.86 & \textbf{86.19} & 81.38 & \textbf{77.74} &  77.30 \\ 
Data Actors       & 76.57  & \textbf{77.87} & \textbf{77.44} & 75.23 & \textbf{76.13} & 72.19  \\ 
Model Owners      & \textbf{96.64} & 95.97 & \textbf{97.77} & 96.05 & \textbf{96.60} & 96.38  \\ 
Model Means       & 64.11  & \textbf{71.16} & 56.21 & \textbf{65.79} & \textbf{69.84} & 68.99  \\ 
App. Means & 50.92 & \textbf{55.14} & 46.00 & \textbf{50.90} & \textbf{66.18} & 60.09  \\ 
Ends              & \textbf{76.58} & 65.26 & \textbf{78.96} & 66.08 & \textbf{74.04} & 59.51  \\ \midrule
\textbf{Total}             & \textbf{73.10} & 73.04 & \textbf{73.76} & 72.57 & \textbf{76.76} & 72.41  \\
\bottomrule
\end{tabular}}
\caption{fMRR for epistemic elements on AFC and HS data. \textit{All} uses all paragraphs from abstracts and introductions; \textit{Oracle} uses paragraphs annotated by humans as containing research framing information.}
\label{tab:filtered_mrr_combined}
\end{table}
We evaluate our semantic clustering-based approach for estimating epistemic element likelihoods in AFC and HS papers. 
Since our approach uses generation rather than classification, the model often produces semantically similar but syntactically varied outputs (e.g., ``not mentioned,” or “absent” when the prediction should be ``not specified''). Without alignment, these would be treated as distinct labels. 
NLI-based clustering addresses this by grouping semantically equivalent outputs—thus supporting effective uncertainty quantification across multiple generations. As shown in Appendix~\ref{semantic-clustering-val}, performance improves with more samples, highlighting the benefits of our approach over a single-prompt baseline.

Table~\ref{tab:filtered_mrr_combined} shows that using all paragraphs slightly outperforms human-filtered ones, suggesting broader retrieval enhances epistemic element extraction, especially given the model’s strong ability to filter irrelevant content (\textasciitilde 76\% fMRR for \textit{not specified} labels).  
The model achieves near-perfect performance in identifying \textit{model owners}, largely due to the high prevalence of \textit{not specified} labels—fewer than 10\% of paragraphs explicitly mention one. In contrast, performance on \textit{application means} is lower, likely due to the greater complexity introduced by at least 10 distinct labels within this category, making classification more challenging. Despite these variations, the model performs consistently well across other epistemic elements. 
Performance is also consistent across domains, except for higher accuracy in identifying \textit{application means} in hate speech detection. This may stem from AFC’s greater label diversity (15 vs. 10), which includes both nuanced tasks (e.g., triaging claims) and concrete actions (e.g., automated removal), the latter shared across both domains.
\subsection{Research framing ranking}
\begin{table}[t]
\centering
\resizebox{.85\linewidth}{!}{
\begin{tabular}{@{}lccc@{}}
\toprule
\textbf{Dataset} & \textbf{Approach} & \textbf{AFC} & \textbf{HS} \\ 
\midrule
\multirow{2}{*}{\textbf{Validation}} 
    & System$_{all}$ &  55.51 & - \\ 
    & System$_{oracle}$ & \textbf{65.04} & - \\ 
\midrule
\multirow{2}{*}{\textbf{Test}} 
    & System$_{all}$ & \textbf{52.10} & \textbf{52.17} \\ 
    & System$_{oracle}$ & 51.93 & 50.50 \\ 
\midrule
\textbf{Entire Dataset} 
    & Human & 84.84 & 83.63 \\ 
\bottomrule
\end{tabular}}
\caption{fMRR for research framing prediction across AFC and HS papers. System$_{all}$ aggregates epistemic element scores from all paragraphs, while System$_{oracle}$ uses only those labeled by humans as containing research framing information for ranking.  }
\label{tab:filtered_mrr_results}
\end{table}
Table \ref{tab:filtered_mrr_results} presents fMRR scores for research framing prediction across AFC and HS datasets.
Despite the complexity of the task—10+ distinct research framings with multiple per paper—both approaches perform well. The upper-bound human performance (84.84\% for AFC, 83.63\% for HS), obtained by applying logical rules to manually annotated epistemic elements, 
indicates that logical mappings successfully capture relationships between epistemic elements and research framings.
%
\subsection{Research framing classification}
Table~\ref{tab:performance_comparison} presents research framing classification results. The baseline LLM, provided with in-context examples, research framing definitions, and paper context, achieves Jaccard scores of 21.89\% (validation) and 28.45\% (test) for AFC, and 16.79\% for HS. This highlights the task’s complexity, as it requires synthesizing multiple epistemic elements and reasoning over their connections.
Augmenting the LLM with ranking-based explanations significantly improves performance, yielding gains of 6–7\% in AFC 
and nearly 15\% in HS 
. This suggests that structured guidance through deterministic research framing ranking explanations enhances inference, particularly in hate speech detection.
\begin{table}[H]
    \centering
    \resizebox{\linewidth}{!}{
    \begin{tabular}{lcc|c}
        \toprule
        & \multicolumn{2}{c}{\textbf{AFC}}  & \textbf{HS} \\
        & \textbf{Validation} & \textbf{Test} & \\
        \midrule
        Baseline & 21.89 $\pm$ 2.08 & 28.45 $\pm$ 1.40 & 16.79 $\pm$ 1.27 \\
        System   & \textbf{30.67} $\pm$ 1.44 & \textbf{34.86} $\pm$ 1.05 & \textbf{31.37} $\pm$ 1.70 \\
        \bottomrule
    \end{tabular}}
    \caption{Jaccard score for research framing classification with 95\% confidence intervals over 15 runs. \textit{Baseline} uses an LLM with in-context examples, research framing definitions, and context, while \textit{System} builds on this by incorporating rankings from our pipeline.}
    \label{tab:performance_comparison}
\end{table}
\begin{table*}[ht!]
    \centering
    \resizebox{\linewidth}{!}{
    \begin{tabular}{p{3.5cm} p{3cm} c c c c | c c c c}
        \toprule
        \textbf{General Framing} & \textbf{Domain Framing} & \multicolumn{4}{c|}{\textbf{AFC}} & \multicolumn{4}{c}{\textbf{HS}} \\
        \cmidrule{3-6} \cmidrule{7-10}
& & 
\textbf{\makecell{C}} & 
\textbf{\makecell{fMRR$_{\text{sys}}$\\(Step 2)}} & 
\textbf{\makecell{F1$_{\text{base}}$\\(Step 3)}} & 
\textbf{\makecell{F1$_{\text{sys}}$\\(Step 3)}} & 
\textbf{\makecell{C}} & 
\textbf{\makecell{fMRR$_{\text{sys}}$\\(Step 2)}} & 
\textbf{\makecell{F1$_{\text{base}}$\\(Step 3)}} & 
\textbf{\makecell{F1$_{\text{sys}}$\\(Step 3)}} \\

        \midrule
        \multirow{2}{*}{Automated deployment} 
            & Automated external fact-checking & 17 & 41.56 & 35.14 $\pm$ 2.70 & \textbf{50.50} $\pm$ 2.28 & -- & -- & -- & -- \\
            & Automated content moderation &  -- & -- & -- & -- &  5 & 51.67 & 10.27 $\pm$ 7.01 & \textbf{27.97} $\pm$ 9.41 \\
        \multirow{2}{*}{Assistive deployment} 
            & Assisted external fact-checking & 17 & 35.93 & 44.11 $\pm$ 2.12 & \textbf{56.79} $\pm$ 2.34 & -- & -- & -- & -- \\
            & Assisted internal fact-checking & 2 & 56.25 & 12.22 $\pm$ 12.94 & 26.48 $\pm$ 10.99 & -- & -- & -- & -- \\
            & Assisted content moderation & -- & -- & -- & -- & 4 & 100.00 & 19.71 $\pm$ 1.31 & \textbf{24.99} $\pm$ 2.84 \\
        \multirow{1}{*}{Knowledge curation}
            & Assisted knowledge curation & 6 & 91.67 & 40.21 $\pm$ 4.73 & \textbf{62.17} $\pm$ 3.46 & -- & -- & -- & -- \\
        \multirow{2}{*}{Knowledge exploration}
            & Scientific curiosity & 5 & 52.41 & 18.92 $\pm$ 4.26 & \textbf{20.24} $\pm$ 2.53 & 16 & 47.43 & 34.60 $\pm$ 3.62 & \textbf{48.84} $\pm$ 2.96 \\
            & Adversarial research & 1 & 50.00 & 46.00 $\pm$ 6.09 & \textbf{72.67} $\pm$ 10.95 & -- & -- & -- & -- \\
        Governance
            & Law enforcement & 1 & 100.00 & 100.00 $\pm$ 0.00 & 100.00 $\pm$ 0.00 & -- & -- & -- & -- \\
        \multirow{3}{*}{Other} 
            & Assisted media consumption & 4 & 36.58 & \textbf{26.23} $\pm$ 4.50 & 18.92 $\pm$ 3.02 & -- & -- & -- & -- \\
            & Automated content moderation & 1 & 14.29 & \textbf{18.16} $\pm$ 6.93 & 0.00 $\pm$ 0.00 & -- & -- & -- & --  \\
            & LLM safety & -- & -- & -- & -- & 2 & 100.00 & 55.56 $\pm$ 4.12 & \textbf{63.11} $\pm$ 7.53 \\
\midrule
          \multirow{2}{*}{Vague deployment}  & Vague debunking & 11 & 71.27 & 18.17 $\pm$ 5.89 & \textbf{31.86} $\pm$ 1.63 & -- & -- & -- & -- \\
            & Vague moderation & -- & -- & -- & -- & 7 & 41.90 & 3.64 $\pm$ 3.81 & \textbf{14.80} $\pm$ 5.69 \\
        Vague identification
            & Vague identification & 25 & 54.82 & 23.86 $\pm$ 4.64 & \textbf{44.97} $\pm$ 2.01 & 34 & 51.30 & 2.92 $\pm$ 2.85 & \textbf{35.79} $\pm$ 1.64 \\
          Vague opposition  & Vague opposition & 13 & 21.50 & 2.23 $\pm$ 2.34 & \textbf{4.35} $\pm$ 3.23 & 0 & -- & -- & -- \\

         Other   & Vague data analysis & -- & -- & -- & -- & 8 & 24.94 & \textbf{61.53} $\pm$ 3.35 & 44.91 $\pm$ 4.03 \\
        \midrule
        \makecell[l]{Underspecified \\ framing}
            &  & 49 & -- & 34.70 $\pm$ 4.07 & \textbf{58.91} $\pm$ 2.21 & 49 & -- & 63.15 $\pm$ 2.50 & \textbf{69.93} $\pm$ 2.70 \\
        \bottomrule
    \end{tabular}}
    \caption{Research framing ranking and classification results in AFC and HS. \textbf{C}: Count. \textbf{fMRR$_{sys}$:} fMRR reporting system performance for ranking (step 2). \textbf{F1$_{\text{base}}$}: classification with baseline LLM. \textbf{F1$_{\text{sys}}$}: classification with baseline LLM incorporating rule-based rankings.  95\% confidence intervals are from 15 runs.}
    \label{tab:framing_results_grouped}
\end{table*}

Table~\ref{tab:framing_results_grouped} reports per-research framing F-scores alongside ranking scores from Step 2. Performance gains strongly correlate with fMRR—higher ranking accuracy leads to greater classification. 
Only two research framings with sufficient samples show declines: \textit{Assisted Media Consumption} (AFC) and \textit{Vague Data Analysis} (HS). A closer examination 
reveals that deterministic rules struggle with context-dependent research framings like \textit{Vague Data Analysis}, which require reasoning about missing connections between means and ends. The LLM mitigates this by integrating context and reasoning to differentiate overlapping research framings.
Nonetheless, most research framings benefit significantly from ranking-based explanations, with \textit{Vague Identification} improving by over 20\% in AFC and over 30\% in HS. Clear research framings (top) 
also show important gains.

To evaluate the impact of confidence levels in research framing ranking justifications on the LLM's final classification, we analyze research framing retention rates. We find that 85.26\% of research framings with high likelihoods (-1 to 0) are included in the final predictions, while 44.64\% of medium-likelihood research framings are retained. In contrast, only 8.47\% of the low-likelihood research framings are kept. This suggests that while the LLM considers confidence scores, it does not rely on them entirely and applies its own reasoning in the final classification.
Finally, the system excels at detecting underspecified research framings, achieving F-scores of 58.91\% (AFC) and 69.93\% (HS) with precision rates of 76.16\% and 86.24\%.

\section{Automated analysis of AFC literature}
\label{sec:meta-analysis}
We apply our system to 102 automated fact-checking papers published between January 2023 and December 2024, 
to identify emerging trends (see Appendix~\ref{app:afc-lit-dist} for distribution details).

Compared to \citet{intendeduses}, we observe a rise in \textit{Scientific Curiosity} framings, driven by the growth of papers analyzing LLM capabilities in fact-checking. The model classifies works like \citet{pinocchio}—focused on evaluating LLM factual reasoning—as scientific curiosity, noting: ``\textit{The primary goal is to understand LLM limitations, a purely scientific endeavor.}'' Papers introducing datasets or benchmarks are similarly grouped under \textit{Scientific Curiosity}, reflecting an emphasis on knowledge development over deployment. 

We also observe a rise in \textit{Vague Debunking} framings and a decline in \textit{Vague Opposition}. The former describes papers that propose automated debunking without clarifying whether models support or replace human fact-checkers; the latter presents a broad goal without specifying how model outputs are used to achieve it. While our system reliably detects underspecification, it sometimes defaults to \textit{Vague Debunking} due to its broader scope—e.g., in \citet{dialogue},  it correctly identifies ambiguity in deployment but infers intended use based on model design, leading to misclassification. 

Furthermore, we observe a shift from fully automated fact-checking (22.0\% in \citet{intendeduses} vs. 7.8\% in our study) toward human-assisted systems (18.0\% vs. 27.5\%), reflecting growing awareness of LLM limitations. 
At the epistemic element level, \textit{supplant human fact-checkers} drops sharply (24.0\% → 3.9\%), while \textit{identifying misleading claims} for human review rises (36.0\% → 55.9\%). 
More papers aim to \textit{generate justifications} (9.0\% → 21.6\%), supporting the trend of AI as assistive, not autonomous. 

\section{Conclusion}
In this work, we first proposed a generalizable research framing schema to support consistent interpretation of research intent across NLP domains, then introduced an automated framework for extracting epistemic elements and inferring framings. To assess generalization, we annotated 49 HS papers and found our system outperforms a strong in-context LLM baseline in both AFC and HS. We applied the framework to 102 recent AFC papers to demonstrate its use for scalable literature analysis—revealing a shift toward human-assisted systems, a rise in underspecified framings, and greater emphasis on justification generation. 
Future work will explore automated framing and epistemic element discovery in new domains to further streamline NLP research analysis.

\section*{Acknowledgments}
Eric Chamoun is supported by an EPSRC-funded
studentship. Andreas Vlachos is supported by the
ERC grant AVeriTeC (GA 865958) and the DARPA program SciFy. Michael Schlichtkrull is supported by the Engineering and Physical Sciences Research Council [grant number EP/Y009800/1], through funding from Responsible AI UK (KP0016). 
We'd like to thank Jose Camacho-Collados for feedback on this paper.
\section*{Limitations}
While our study introduces an automated framework for meta-analyses in NLP research, we acknowledge the following:

\begin{itemize}
    \item \textbf{Scope of defined research framings} – The predefined research framings, derived manually by domain experts from several papers, provide a strong foundation. However, they may not fully capture the evolving perspectives and emerging trends within the field.
    
    \item \textbf{Generalizability of conclusions} – Our findings reveal important trends, such as the prevalence of underspecified research framings. However, given that our analysis does not cover all NLP papers on automated fact-checking and hate speech detection, these insights should be viewed as well-informed estimates—not definitive claims. Nonetheless, they offer a strong starting point for understanding broader trends in research framing.
    
    \item \textbf{Tool accuracy} – While our model is effective at identifying underspecified research framings, achieving a precision of 76.16\% in fact-checking and 86.24\% in hate speech detection, we acknowledge that, as any other automated framework, it still makes mistakes. Individual misclassifications can occur, though large-scale trends remain informative. 
    
    \item \textbf{Appropriate use cases} – The tool is designed primarily for scalable analyses of new subfields (where small errors cancel out in aggregate analyses) and \textit{writing assistance} (offering feedback for authors on the writing, akin to systems such as SWIF$^{2}$T \citep{swift}). It is \textbf{not intended for peer review or paper acceptance decisions}, as it is not precise enough to make definitive judgments about research quality. Furthermore, our aim is \textbf{not} to prescribe how research should be framed, but to surface and clarify existing framings—whether practical, exploratory, or societal—to support reflection on how NLP contributions are positioned and interpreted.
    
    \item \textbf{Human oversight} – Automation enhances scalability but does \textbf{not} replace expert judgment. The tool should be used as a \textit{supplementary resource}, requiring human review in critical applications.
\end{itemize}

Despite these limitations, our approach provides valuable insights into NLP research trends and can serve as a promising writing assistant, supporting both large-scale analysis and research framing refinement.

\section*{Ethical considerations}
Our study analyzes publicly available NLP research papers to examine how they frame the intended use of their artefacts. All data used, including annotated excerpts, is drawn from these publications and included in our supplementary materials. Since we focus on textual analysis of already public documents, we do not anticipate ethical concerns. However, if any author requests that specific excerpts from their work be removed from our dataset or repository, we will comply immediately.

\newpage 
\bibliography{custom}

\begin{thebibliography}{23}
\providecommand{\natexlab}[1]{#1}

\bibitem[{Bender et~al.(2021)Bender, Gebru, McMillan-Major, and Shmitchell}]{bender}
Emily~M. Bender, Timnit Gebru, Angelina McMillan-Major, and Shmargaret Shmitchell. 2021.
\newblock \href {https://doi.org/10.1145/3442188.3445922} {On the dangers of stochastic parrots: Can language models be too big?}
\newblock In \emph{Proceedings of the 2021 ACM Conference on Fairness, Accountability, and Transparency}, FAccT '21, page 610–623, New York, NY, USA. Association for Computing Machinery.

\bibitem[{Bian et~al.(2020)Bian, Xiao, Xu, Zhao, Huang, Rong, and Huang}]{fc-example}
Tian Bian, Xi~Xiao, Tingyang Xu, Peilin Zhao, Wenbing Huang, Yu~Rong, and Junzhou Huang. 2020.
\newblock \href {https://doi.org/10.1609/aaai.v34i01.5393} {Rumor detection on social media with bi-directional graph convolutional networks}.
\newblock \emph{Proceedings of the AAAI Conference on Artificial Intelligence}, 34(01):549--556.

\bibitem[{Birhane et~al.(2022)Birhane, Kalluri, Card, Agnew, Dotan, and Bao}]{birhane}
Abeba Birhane, Pratyusha Kalluri, Dallas Card, William Agnew, Ravit Dotan, and Michelle Bao. 2022.
\newblock \href {https://doi.org/10.1145/3531146.3533083} {The values encoded in machine learning research}.
\newblock In \emph{Proceedings of the 2022 ACM Conference on Fairness, Accountability, and Transparency}, FAccT '22, page 173–184, New York, NY, USA. Association for Computing Machinery.

\bibitem[{Bordes et~al.(2013)Bordes, Usunier, Garcia-Duran, Weston, and Yakhnenko}]{fmrr}
Antoine Bordes, Nicolas Usunier, Alberto Garcia-Duran, Jason Weston, and Oksana Yakhnenko. 2013.
\newblock \href {https://proceedings.neurips.cc/paper_files/paper/2013/file/1cecc7a77928ca8133fa24680a88d2f9-Paper.pdf} {Translating embeddings for modeling multi-relational data}.
\newblock In \emph{Advances in Neural Information Processing Systems}, volume~26. Curran Associates, Inc.

\bibitem[{Breiman et~al.(1984)Breiman, Friedman, Olshen, and Stone}]{cart}
L.~Breiman, Jerome~H. Friedman, Richard~A. Olshen, and C.~J. Stone. 1984.
\newblock \href {https://api.semanticscholar.org/CorpusID:29458883} {Classification and regression trees}.
\newblock \emph{Biometrics}, 40:874.

\bibitem[{Card et~al.(2016)Card, Gross, Boydstun, and Smith}]{card-etal-2016-analyzing}
Dallas Card, Justin Gross, Amber Boydstun, and Noah~A. Smith. 2016.
\newblock \href {https://doi.org/10.18653/v1/D16-1148} {Analyzing framing through the casts of characters in the news}.
\newblock In \emph{Proceedings of the 2016 Conference on Empirical Methods in Natural Language Processing}, pages 1410--1420, Austin, Texas. Association for Computational Linguistics.

\bibitem[{Chamoun et~al.(2023)Chamoun, Saeidi, and Vlachos}]{dialogue}
Eric Chamoun, Marzieh Saeidi, and Andreas Vlachos. 2023.
\newblock \href {https://doi.org/10.18653/v1/2023.emnlp-main.993} {Automated fact-checking in dialogue: Are specialized models needed?}
\newblock In \emph{Proceedings of the 2023 Conference on Empirical Methods in Natural Language Processing}, pages 16009--16020, Singapore. Association for Computational Linguistics.

\bibitem[{Chamoun et~al.(2024)Chamoun, Schlichtkrull, and Vlachos}]{swift}
Eric Chamoun, Michael Schlichtkrull, and Andreas Vlachos. 2024.
\newblock \href {https://doi.org/10.18653/v1/2024.findings-acl.580} {Automated focused feedback generation for scientific writing assistance}.
\newblock In \emph{Findings of the Association for Computational Linguistics: ACL 2024}, pages 9742--9763, Bangkok, Thailand. Association for Computational Linguistics.

\bibitem[{{Gemini Team} et~al.(2023){Gemini Team}, {Anil}, {Borgeaud}, {Alayrac}, {Yu}, {Soricut}, {Schalkwyk}, {Dai}, {Hauth}, {Millican}, {Silver}, {Johnson}, {Antonoglou}, {Schrittwieser}, {Glaese}, {Chen}, {Pitler}, {Lillicrap}, {Lazaridou}, {Firat}, {Molloy}, {Isard}, {Barham}, {Hennigan}, {Lee}, {Viola}, {Reynolds}, {Xu}, {Doherty}, {Collins}, {Meyer}, {Rutherford}, {Moreira}, {Ayoub}, {Goel}, {Krawczyk}, {Du}, {Chi}, {Cheng}, {Ni}, {Shah}, {Kane}, {Chan}, {Faruqui}, {Severyn}, {Lin}, {Li}, {Cheng}, {Ittycheriah}, {Mahdieh}, {Chen}, {Sun}, {Tran}, {Bagri}, {Lakshminarayanan}, {Liu}, {Orban}, {G{\"u}ra}, {Zhou}, {Song}, {Boffy}, {Ganapathy}, {Zheng}, {Choe}, {Weisz}, {Zhu}, {Lu}, {Gopal}, {Kahn}, {Kula}, {Pitman}, {Shah}, {Taropa}, {Al Merey}, {Baeuml}, {Chen}, {El Shafey}, {Zhang}, {Sercinoglu}, {Tucker}, {Piqueras}, {Krikun}, {Barr}, {Savinov}, {Danihelka}, {Roelofs}, {White}, {Andreassen}, {von Glehn}, {Yagati}, {Kazemi}, {Gonzalez}, {Khalman}, {Sygnowski}, {Frechette}, {Smith}, {Culp}, {Proleev},
  {Luan}, {Chen}, {Lottes}, {Schucher}, {Lebron}, {Rrustemi}, {Clay}, {Crone}, {Kocisky}, {Zhao}, {Perz}, {Yu}, {Howard}, {Bloniarz}, {Rae}, {Lu}, {Sifre}, {Maggioni}, {Alcober}, {Garrette}, {Barnes}, {Thakoor}, {Austin}, {Barth-Maron}, {Wong}, {Joshi}, {Chaabouni}, {Fatiha}, {Ahuja}, {Singh Tomar}, {Senter}, {Chadwick}, {Kornakov}, {Attaluri}, {Iturrate}, {Liu}, {Li}, {Cogan}, {Chen}, {Jia}, {Gu}, {Zhang}, {Grimstad}, {Jakse Hartman}, {Garcia}, {Sankaranarayana Pillai}, {Devlin}, {Laskin}, {de Las Casas}, {Valter}, {Tao}, {Blanco}, {Puigdom{\`e}nech Badia}, {Reitter}, {Chen}, {Brennan}, {Rivera}, {Brin}, {Iqbal}, {Surita}, {Labanowski}, {Rao}, {Winkler}, {Parisotto}, {Gu}, {Olszewska}, {Addanki}, {Miech}, {Louis}, {Teplyashin}, {Brown}, {Catt}, {Balaguer}, {Xiang}, {Wang}, {Ashwood}, {Briukhov}, {Webson}, {Ganapathy}, {Sanghavi}, {Kannan}, {Chang}, {Stjerngren}, {Djolonga}, {Sun}, {Bapna}, {Aitchison}, {Pejman}, {Michalewski}, {Yu}, {Wang}, {Love}, {Ahn}, {Bloxwich}, {Han}, {Humphreys}, {Sellam}, {Bradbury},
  {Godbole}, {Samangooei}, {Damoc}, and {Kaskasoli}}]{gemini}
{Gemini Team}, Rohan {Anil}, Sebastian {Borgeaud}, Jean-Baptiste {Alayrac}, Jiahui {Yu}, Radu {Soricut}, Johan {Schalkwyk}, Andrew~M. {Dai}, Anja {Hauth}, Katie {Millican}, David {Silver}, Melvin {Johnson}, Ioannis {Antonoglou}, Julian {Schrittwieser}, Amelia {Glaese}, Jilin {Chen}, Emily {Pitler}, Timothy {Lillicrap}, Angeliki {Lazaridou}, Orhan {Firat}, James {Molloy}, Michael {Isard}, Paul~R. {Barham}, Tom {Hennigan}, Benjamin {Lee}, Fabio {Viola}, Malcolm {Reynolds}, Yuanzhong {Xu}, Ryan {Doherty}, Eli {Collins}, Clemens {Meyer}, Eliza {Rutherford}, Erica {Moreira}, Kareem {Ayoub}, Megha {Goel}, Jack {Krawczyk}, Cosmo {Du}, Ed~{Chi}, Heng-Tze {Cheng}, Eric {Ni}, Purvi {Shah}, Patrick {Kane}, Betty {Chan}, Manaal {Faruqui}, Aliaksei {Severyn}, Hanzhao {Lin}, YaGuang {Li}, Yong {Cheng}, Abe {Ittycheriah}, Mahdis {Mahdieh}, Mia {Chen}, Pei {Sun}, Dustin {Tran}, Sumit {Bagri}, Balaji {Lakshminarayanan}, Jeremiah {Liu}, Andras {Orban}, Fabian {G{\"u}ra}, Hao {Zhou}, Xinying {Song}, Aurelien {Boffy}, Harish
  {Ganapathy}, Steven {Zheng}, HyunJeong {Choe}, {\'A}goston {Weisz}, Tao {Zhu}, Yifeng {Lu}, Siddharth {Gopal}, Jarrod {Kahn}, Maciej {Kula}, Jeff {Pitman}, Rushin {Shah}, Emanuel {Taropa}, Majd {Al Merey}, Martin {Baeuml}, Zhifeng {Chen}, Laurent {El Shafey}, Yujing {Zhang}, Olcan {Sercinoglu}, George {Tucker}, Enrique {Piqueras}, Maxim {Krikun}, Iain {Barr}, Nikolay {Savinov}, Ivo {Danihelka}, Becca {Roelofs}, Ana{\"\i}s {White}, Anders {Andreassen}, Tamara {von Glehn}, Lakshman {Yagati}, Mehran {Kazemi}, Lucas {Gonzalez}, Misha {Khalman}, Jakub {Sygnowski}, Alexandre {Frechette}, Charlotte {Smith}, Laura {Culp}, Lev {Proleev}, Yi~{Luan}, Xi~{Chen}, James {Lottes}, Nathan {Schucher}, Federico {Lebron}, Alban {Rrustemi}, Natalie {Clay}, Phil {Crone}, Tomas {Kocisky}, Jeffrey {Zhao}, Bartek {Perz}, Dian {Yu}, Heidi {Howard}, Adam {Bloniarz}, Jack~W. {Rae}, Han {Lu}, Laurent {Sifre}, Marcello {Maggioni}, Fred {Alcober}, Dan {Garrette}, Megan {Barnes}, Shantanu {Thakoor}, Jacob {Austin}, Gabriel {Barth-Maron},
  William {Wong}, Rishabh {Joshi}, Rahma {Chaabouni}, Deeni {Fatiha}, Arun {Ahuja}, Gaurav {Singh Tomar}, Evan {Senter}, Martin {Chadwick}, Ilya {Kornakov}, Nithya {Attaluri}, I{\~n}aki {Iturrate}, Ruibo {Liu}, Yunxuan {Li}, Sarah {Cogan}, Jeremy {Chen}, Chao {Jia}, Chenjie {Gu}, Qiao {Zhang}, Jordan {Grimstad}, Ale {Jakse Hartman}, Xavier {Garcia}, Thanumalayan {Sankaranarayana Pillai}, Jacob {Devlin}, Michael {Laskin}, Diego {de Las Casas}, Dasha {Valter}, Connie {Tao}, Lorenzo {Blanco}, Adri{\`a} {Puigdom{\`e}nech Badia}, David {Reitter}, Mianna {Chen}, Jenny {Brennan}, Clara {Rivera}, Sergey {Brin}, Shariq {Iqbal}, Gabriela {Surita}, Jane {Labanowski}, Abhi {Rao}, Stephanie {Winkler}, Emilio {Parisotto}, Yiming {Gu}, Kate {Olszewska}, Ravi {Addanki}, Antoine {Miech}, Annie {Louis}, Denis {Teplyashin}, Geoff {Brown}, Elliot {Catt}, Jan {Balaguer}, Jackie {Xiang}, Pidong {Wang}, Zoe {Ashwood}, Anton {Briukhov}, Albert {Webson}, Sanjay {Ganapathy}, Smit {Sanghavi}, Ajay {Kannan}, Ming-Wei {Chang}, Axel
  {Stjerngren}, Josip {Djolonga}, Yuting {Sun}, Ankur {Bapna}, Matthew {Aitchison}, Pedram {Pejman}, Henryk {Michalewski}, Tianhe {Yu}, Cindy {Wang}, Juliette {Love}, Junwhan {Ahn}, Dawn {Bloxwich}, Kehang {Han}, Peter {Humphreys}, Thibault {Sellam}, James {Bradbury}, Varun {Godbole}, Sina {Samangooei}, Bogdan {Damoc}, and Alex {Kaskasoli}. 2023.
\newblock \href {https://doi.org/10.48550/arXiv.2312.11805} {{Gemini: A Family of Highly Capable Multimodal Models}}.
\newblock \emph{arXiv e-prints}, arXiv:2312.11805.

\bibitem[{Gooding et~al.(2025)Gooding, Lopez-Rivilla, and Grefenstette}]{writingastestbed}
Sian Gooding, Lucia Lopez-Rivilla, and Edward Grefenstette. 2025.
\newblock \href {https://arxiv.org/abs/2503.19711} {Writing as a testbed for open ended agents}.
\newblock \emph{Preprint}, arXiv:2503.19711.

\bibitem[{Grodzinsky et~al.(2012)Grodzinsky, Miller, and Wolf}]{grodzinsky}
F.~S. Grodzinsky, K.~Miller, and M.~J. Wolf. 2012.
\newblock \href {https://doi.org/10.1145/2422509.2422511} {Moral responsibility for computing artifacts: "the rules" and issues of trust}.
\newblock \emph{SIGCAS Comput. Soc.}, 42(2):15–25.

\bibitem[{He et~al.(2020)He, Liu, Gao, and Chen}]{deberta}
Pengcheng He, Xiaodong Liu, Jianfeng Gao, and Weizhu Chen. 2020.
\newblock \href {https://arxiv.org/abs/2006.03654} {Deberta: Decoding-enhanced {BERT} with disentangled attention}.
\newblock \emph{CoRR}, abs/2006.03654.

\bibitem[{Hu et~al.(2024)Hu, Chen, Li, Guo, Wen, Yu, and Guo}]{pinocchio}
Xuming Hu, Junzhe Chen, Xiaochuan Li, Yufei Guo, Lijie Wen, Philip~S. Yu, and Zhijiang Guo. 2024.
\newblock \href {https://doi.org/10.48550/ARXIV.2310.05177} {Towards understanding factual knowledge of large language models}.
\newblock \emph{12th International Conference on Learning Representations, {ICLR} 2024, Messe Wien Exhibition and Congress Center, Vienna Austria May 7th, 2024 to May 11th, 2024}, 2024.

\bibitem[{Jaccard(1901)}]{jaccard}
Paul Jaccard. 1901.
\newblock {\'E}tude comparative de la distribution florale dans une portion des alpes et des jura.
\newblock \emph{Bull Soc Vaudoise Sci Nat}, 37:547--579.

\bibitem[{Jin et~al.(2021)Jin, Chauhan, Tse, Sachan, and Mihalcea}]{nlp4sg}
Zhijing Jin, Geeticka Chauhan, Brian Tse, Mrinmaya Sachan, and Rada Mihalcea. 2021.
\newblock \href {https://doi.org/10.18653/v1/2021.findings-acl.273} {How good is {NLP}? a sober look at {NLP} tasks through the lens of social impact}.
\newblock In \emph{Findings of the Association for Computational Linguistics: ACL-IJCNLP 2021}, pages 3099--3113, Online. Association for Computational Linguistics.

\bibitem[{Konstantinovskiy et~al.(2021)Konstantinovskiy, Price, Babakar, and Zubiaga}]{fc-example-new}
Lev Konstantinovskiy, Oliver Price, Mevan Babakar, and Arkaitz Zubiaga. 2021.
\newblock \href {https://doi.org/10.1145/3412869} {Toward automated factchecking: Developing an annotation schema and benchmark for consistent automated claim detection}.
\newblock \emph{Digital Threats}, 2(2).

\bibitem[{Krippendorff(2018)}]{krippendorff}
Klaus Krippendorff. 2018.
\newblock \emph{Content analysis: An introduction to its methodology}.
\newblock Sage publications.

\bibitem[{Kuhn et~al.(2023)Kuhn, Gal, and Farquhar}]{semantic-entropy}
Lorenz Kuhn, Yarin Gal, and Sebastian Farquhar. 2023.
\newblock \href {https://arxiv.org/abs/2302.09664} {Semantic uncertainty: Linguistic invariances for uncertainty estimation in natural language generation}.
\newblock \emph{Preprint}, arXiv:2302.09664.

\bibitem[{Leins et~al.(2020)Leins, Lau, and Baldwin}]{leins2020}
Kobi Leins, Jey~Han Lau, and Timothy Baldwin. 2020.
\newblock \href {https://doi.org/10.18653/v1/2020.acl-main.261} {Give me convenience and give her death: Who should decide what uses of {NLP} are appropriate, and on what basis?}
\newblock In \emph{Proceedings of the 58th Annual Meeting of the Association for Computational Linguistics}, pages 2908--2913, Online. Association for Computational Linguistics.

\bibitem[{Liu et~al.(2023)Liu, Cao, Blodgett, Cheung, Olteanu, and Trischler}]{liu-etal-2023-responsible}
Yu~Lu Liu, Meng Cao, Su~Lin Blodgett, Jackie Chi~Kit Cheung, Alexandra Olteanu, and Adam Trischler. 2023.
\newblock \href {https://doi.org/10.18653/v1/2023.findings-emnlp.413} {Responsible {AI} considerations in text summarization research: A review of current practices}.
\newblock In \emph{Findings of the Association for Computational Linguistics: EMNLP 2023}, pages 6246--6261, Singapore. Association for Computational Linguistics.

\bibitem[{Noble(2018)}]{noble}
Safiya~Umoja Noble. 2018.
\newblock \href {http://www.jstor.org/stable/j.ctt1pwt9w5} {\emph{Algorithms of Oppression: How Search Engines Reinforce Racism}}.
\newblock NYU Press.

\bibitem[{Ousidhoum et~al.(2019)Ousidhoum, Lin, Zhang, Song, and Yeung}]{hatespeech-example}
Nedjma Ousidhoum, Zizheng Lin, Hongming Zhang, Yangqiu Song, and Dit-Yan Yeung. 2019.
\newblock \href {https://doi.org/10.18653/v1/D19-1474} {Multilingual and multi-aspect hate speech analysis}.
\newblock In \emph{Proceedings of the 2019 Conference on Empirical Methods in Natural Language Processing and the 9th International Joint Conference on Natural Language Processing (EMNLP-IJCNLP)}, pages 4675--4684, Hong Kong, China. Association for Computational Linguistics.

\bibitem[{Schlichtkrull et~al.(2023)Schlichtkrull, Ousidhoum, and Vlachos}]{intendeduses}
Michael Schlichtkrull, Nedjma Ousidhoum, and Andreas Vlachos. 2023.
\newblock \href {https://doi.org/10.18653/v1/2023.findings-emnlp.577} {The intended uses of automated fact-checking artefacts: Why, how and who}.
\newblock In \emph{Findings of the Association for Computational Linguistics: EMNLP 2023}, pages 8618--8642, Singapore. Association for Computational Linguistics.

\end{thebibliography}
\clearpage
\appendix
\section{List of Epistemic Elements and Research Framings}
\label{appendix:universe}
For clarity and completeness, we list below the full set of canonical labels used in epistemic element extraction and research framing classification. For AFC, these labels are drawn from \citet{intendeduses}; full definitions can be found in their original paper. For HS, we introduce a new set of canonical labels, defined specifically for this study. Detailed descriptions of each label are provided in Appendix~\ref{annotation-appendix}.

\subsection{Fact-checking}
\label{appendix:universe-fc}
\paragraph{Epistemic Elements}

\begin{itemize}
    \item \textbf{Data Subjects}:\\
    \textit{professional journalists}, \textit{citizen journalists}, \textit{social media users}, \textit{technical writers}, \textit{public figures/politicians}, \textit{product reviewers}, \textit{suspects}
    
    \item \textbf{Data Actors}:\\
    \textit{publishers}, \textit{professional journalists}, \textit{product reviewers}, \textit{citizen journalists}, \textit{scientists}, \textit{media consumers}, \textit{technical writers}, \textit{engineers and curators}, \textit{law enforcement}, \textit{algorithm}
    
    \item \textbf{Model Owners}:\\
    \textit{social media companies}, \textit{law enforcement}, \textit{scientists}, \textit{professional journalists}
    
    \item \textbf{Modeling Means}:\\
    \textit{classify/score veracity}, \textit{classify/score stance}, \textit{evidence retrieval}, \textit{justification/explanation production}, \textit{corpora analysis}, \textit{human in the loop}, \textit{generate claims}, \textit{triage claims}
    
    \item \textbf{Application Means}:\\
    \textit{identify claims}, \textit{triage claims}, \textit{supplant human fact-checkers}, \textit{gather and present evidence}, \textit{identify multimodal inconsistencies}, \textit{automated removal}, \textit{provide labels/veracity scores}, \textit{provide aggregates of social media comments}, \textit{filter system outputs}, \textit{maintain consistency with knowledge base}, \textit{analyse data}, \textit{produce misinformation}, \textit{vague persuasion}
    
    \item \textbf{Epistemic Ends}:\\
    \textit{limit misinformation}, \textit{limit ai-generated misinformation}, \textit{increase veracity of published content}, \textit{develop knowledge of NLP/language}, \textit{avoid biases of human fact-checkers}, \textit{detect falsehood for law enforcement}
\end{itemize}

\paragraph{Research Framings}
\vspace{-1em}
\begin{itemize}
    \item \textit{adversarial research}, \textit{assisted external fact-checking}, \textit{assisted internal fact-checking}, \textit{assisted knowledge curation}, \textit{assisted media consumption}, \textit{automated content moderation}, \textit{automated external fact-checking}, \textit{scientific curiosity}, \textit{truth-telling for law enforcement}, \textit{vague debunking}, \textit{vague identification}, \textit{vague opposition}
\end{itemize}

\subsection{Hate Speech Detection}
\label{appendix:universe-hs}
\paragraph{Epistemic Elements}

\begin{itemize}
    \item \textbf{Data Subjects}:\\
    \textit{journalists}, \textit{social media users}, \textit{public figures/politicians}, \textit{product reviewers}, \textit{moderators}, \textit{media consumers}
    
    \item \textbf{Data Actors}:\\
    \textit{social media moderators}, \textit{social media companies}, \textit{scientists}, \textit{media consumers}, \textit{engineers and curators}, \textit{law enforcement}, \textit{algorithm}
    
    \item \textbf{Model Owners}:\\
    \textit{media companies}, \textit{social media companies}, \textit{law enforcement}
    
    \item \textbf{Modeling Means}:\\
    \textit{classify/score posts/text}, \textit{provide justifications}, \textit{human in the loop}, \textit{corpora analysis}, \textit{data collection}
    
    \item \textbf{Application Means}:\\
    \textit{identify toxic content}, \textit{supplant human moderators}, \textit{gather and present justification}, \textit{automated removal}, \textit{provide labels/toxicity scores}, \textit{filter system outputs}, \textit{analyse data}, \textit{vague persuasion}, \textit{generate counter narratives}
    
    \item \textbf{Epistemic Ends}:\\
    \textit{fight hate}, \textit{counter hate}, \textit{limit ai-generated toxicity/misinformation}, \textit{develop knowledge of NLP/language}, \textit{avoid biases of human moderators}, \textit{detect hate/falsehood for law enforcement}
\end{itemize}

\paragraph{Research Framings}
\vspace{-1em}
\begin{itemize}
    \item \textit{assisted content moderation}, \textit{assisted knowledge curation}, \textit{automatic content moderation}, \textit{LLM safety}, \textit{scientific curiosity}, \textit{truth-telling for law enforcement}, \textit{vague data analysis}, \textit{vague identification}, \textit{vague moderation}, \textit{vague opposition}
\end{itemize}

\clearpage
\onecolumn
\section{Epistemic element extraction}
\subsection{Semantic clustering via NLI validation}
\label{semantic-clustering-val}
\begin{table}[h]
\centering
\label{tab:fmrr-results}
\resizebox{\textwidth}{!}{
\begin{tabular}{c|cccccc|cccccc}
\toprule
\multirow{2}{*}{\textbf{Samples}} & \multicolumn{6}{c|}{\textbf{Automated Fact-Checking}} & \multicolumn{6}{c}{\textbf{Hate Speech}} \\
 & Overall & DS & DA & MO & MM & EN & Overall & DS & DA & MO & MM & EN \\
\midrule
1  & 68.18 & 82.12 & 68.32 & 96.90 & 52.25 & 67.81 & 69.59 & 70.38 & 67.04 & 96.82 & 59.24 & 69.19 \\
3  & 70.17 & 83.21 & 71.11 & 96.90 & 54.76 & 71.44 & 72.34 & 74.12 & 71.82 & 97.13 & 60.09 & 72.05 \\
5  & 71.59 & 84.29 & 72.65 & 97.02 & \textbf{56.61} & 73.93 & 72.52 & 74.68 & 71.84 & 96.82 & 60.84 & 73.11 \\
8  & 72.12 & 84.73 & 73.97 & 97.45 & 56.29 & 73.50 & 73.80 & 73.78 & 72.74 & \textbf{97.77} & 62.09 & \textbf{74.76} \\
10 & \textbf{73.76} & \textbf{86.19} & \textbf{77.44} & \textbf{97.77} & 56.21 & \textbf{78.96} & \textbf{76.76} & \textbf{77.74} & \textbf{76.13} & 96.60 & \textbf{69.84 }& 74.04 \\
\bottomrule
\end{tabular}
}
\caption{Filtered MRR (\%) for each epistemic element across sample sizes. DS = Data Subjects, DA = Data Actors, MO = Model Owners, MM = Model Means, AM = Application Means, EN = Epistemic Ends.}
\end{table}
\section{Research Framing Ranking}
\subsection{Logical mappings}
\label{logical-mappings-appendix}
\subsubsection{Automated fact-checking}
\noindent\textbf{Adversarial Research:}
\begin{align*}
    &\text{Data actors: \textit{scientists}} \\
    &\land \, \text{Model means: \textit{generate claims}} \\
    &\land \, \text{Application means: \textit{produce misinformation}}
\end{align*}

\vspace{0.5em}

\noindent\textbf{Assisted External Fact-checking:}
\begin{align*}
    &\mathbf{\bigvee} \big\{ 
        \text{Data subjects: All except \textit{professional journalists}, \textit{citizen journalists}, \textit{technical writers}} 
    \big\} \\
    &\land \, \mathbf{\bigvee} \big\{ 
        \text{Data actors: \textit{professional journalists}}, 
        \textit{citizen journalists}, 
        \textit{technical writers} 
    \big\} \\
    &\land \, \mathbf{\bigvee} \big\{ 
        \text{Application means: All except \textit{supplant human fact-checkers}, \textit{not specified}} 
    \big\} \\
    &\land \, \mathbf{\bigvee} \big\{ 
        \text{Ends: \textit{limit misinformation}}, 
        \textit{limit AI-generated misinformation} 
    \big\}
\end{align*}

\vspace{0.5em}

\noindent\textbf{Assisted Internal Fact-checking:}
\begin{align*}
    &\mathbf{\bigvee} \big\{ 
        \text{Data subjects: \textit{professional journalists}}, 
        \textit{citizen journalists}, 
        \textit{technical writers} 
    \big\} \\
    &\land \, \mathbf{\bigvee} \big\{ 
        \text{Data actors: \textit{professional journalists}}, 
        \textit{citizen journalists}, 
        \textit{technical writers} 
    \big\} \\
    &\land \, \mathbf{\bigvee} \big\{ 
        \text{Application means: All except \textit{supplant human fact-checkers}, \textit{not specified}} 
    \big\} \\
    &\land \, \mathbf{\bigvee} \big\{ 
        \text{Ends: \textit{increase veracity of published content}}, 
        \textit{limit misinformation} 
    \big\}
\end{align*}

\vspace{0.5em}

\noindent\textbf{Assisted Knowledge Curation:}
\begin{align*}
    &\mathbf{\bigvee} \big\{ 
        \text{Data actors: \textit{engineers and curators}}, 
        \text{Application means: \textit{filter system outputs}} 
    \big\} \\
    &\mathbf{\land} \, \mathbf{\bigvee} \big\{ 
        \text{Model means: \textit{classify/score veracity}}, 
        \textit{evidence retrieval}, \textit{human in the loop}
    \big\} \\
    &\mathbf{\land} \, \text{Ends: \textit{increase veracity of published content}} 
\end{align*}
\vspace{0.5em}

\noindent\textbf{Assisted Media Consumption:}
\[
\mathbf{\bigvee} \left\{
\begin{aligned}
    &\big( \text{Data subjects: \textit{social media users}} \, \mathbf{\land} \, \text{Model owners: \textit{social media companies}} \\
    &\quad \mathbf{\land} \, \mathbf{\bigvee} \big\{ \text{Application means: \textit{vague persuasion}}, \, \text{Ends: \textit{limit misinformation}} \big\} \big), \\
    &\text{Data actors: \textit{media consumers}}
\end{aligned}
\right.
\]
\vspace{0.5em}

\vspace{0.5em}

\noindent\textbf{Automated Content Moderation:}
\[
\mathbf{\bigvee} \left\{
\begin{aligned}
    &\big( \text{Data subjects: \textit{social media users}} \, \mathbf{\land} \,\text{Data actors: \textit{algorithm}} \\
    &\quad \mathbf{\land} \,    \text{Model owners: \textit{social media companies}} \big),  \\
    &\text{Application means: \textit{automated removal}}
\end{aligned}
\right.
\]
\vspace{0.5em}

\noindent\textbf{Automated External Fact-checking:}
\begin{align*}
    &\text{Application means: \textit{supplant human fact-checkers}}
\end{align*}

\vspace{0.5em}

\noindent\textbf{Scientific Curiosity:}
\begin{align*}
    &\Big( \text{Data actors: \textit{scientists}} \land \text{Model owners: \textit{scientists}} \Big) \\
    &\lor \, \text{Ends: \textit{develop knowledge of NLP/language}}
\end{align*}

\vspace{0.5em}

\noindent\textbf{Vague Debunking:}
\begin{align*}
    &\text{Data actors: \textit{not specified}} \\
    &\land \, \mathbf{\bigvee} \big\{ 
        \text{Model means: \textit{evidence retrieval}, \textit{classify/score veracity}}        
    \big\} \\
    &\land \, \mathbf{\bigvee} \big\{ 
        \text{Application means: \textit{gather and present evidence}}, 
        \textit{provide labels/veracity scores} 
    \big\}
\end{align*}
\vspace{0.5em}

\noindent\textbf{Vague Identification:}
\begin{align*}
    &\text{Data actors: \textit{not specified}} \\
    &\land \, \text{Application means: \textit{identify claims}}
\end{align*}

\vspace{0.5em}

\noindent\textbf{Vague Opposition:}
\begin{align*}
    &\text{Application means: \textit{not specified}} \\
    &\land \, \mathbf{\bigvee} \big\{ 
        \text{Ends: \textit{limit misinformation}}, 
        \textit{limit AI-generated misinformation} 
    \big\}
\end{align*}
\newpage
\subsubsection{Hate Speech Detection}

\noindent\textbf{Assisted Content Moderation:}
\begin{align*}
    &\mathbf{\bigvee} \left\{ 
        \begin{aligned}
            &\text{Model means: \textit{human in the loop}}, \\ 
            &\text{Data actors: \textit{social media moderators}} 
        \end{aligned} 
    \right\}
\end{align*}
\vspace{0.5em}

\noindent\textbf{Assisted Knowledge Curation:}
\begin{align*}
    &\text{Application means: \textit{filter system outputs}} \\
    &\land \, \text{Data actors: \textit{engineers and curators}}
\end{align*}

\vspace{0.5em}
\noindent\textbf{Automatic Content Moderation:}
\begin{align*}
    &\mathbf{\bigvee} \left\{ 
        \begin{aligned}
            &\text{Application means: \textit{supplant human moderators}}, \\ 
            &\text{Application means: \textit{automated removal}}, \\ 
            &\text{Application means: \textit{generate counter narratives}}
        \end{aligned} 
    \right\}
\end{align*}

\vspace{0.5em}

\noindent\textbf{LLM Safety:}
\begin{align*}
    &\text{Ends: \textit{limit AI-generated toxicity/misinformation}}
\end{align*}

\vspace{0.5em}

\noindent\textbf{Scientific Curiosity:}
\begin{align*}
\mathbf{\bigvee} \left\{
\begin{aligned}
    &\text{Data actors: \textit{scientists}}, \\ 
    &\text{Application means: \textit{analyze data}}, \\ 
    &\text{Model means: \textit{corpora analysis}} 
\end{aligned}
\right\}
    &\land \, \text{Ends: \textit{develop knowledge of NLP/language}}
\end{align*}

\vspace{0.5em}

\noindent\textbf{Truth-telling for Law Enforcement:}
\begin{align*}
    &\text{Data actors: \textit{law enforcement}} \\
    &\land \, \text{Ends: \textit{detect hate/falsehood for law enforcement}}
\end{align*}

\vspace{0.5em}

\noindent\textbf{Vague Data Analysis:}
\begin{align*}
    &\mathbf{\bigvee} \left\{ 
        \begin{aligned}
            &\text{Application means: \textit{analyze data}}, \\ 
            &\text{Model means: \textit{corpora analysis}}, \\ 
            &\text{Model means: \textit{data collection}} 
        \end{aligned} 
    \right\} \\
    &\land \, \text{Ends: \textit{fight hate}}
\end{align*}

\vspace{0.5em}

\noindent\textbf{Vague Identification:}
\begin{align*}
    &\text{Data actors: \textit{not specified}} \\
    &\land \, \text{Application means: \textit{identify toxic content}}
\end{align*}

\vspace{0.5em}

\noindent\textbf{Vague Moderation:}
\begin{align*}
    &\mathbf{\bigvee} \left\{ 
        \begin{aligned}
            &\text{Data actors: \textit{social media moderators}}, \\ 
            &\text{Data actors: \textit{social media companies}}, \\ 
            &\text{Model owners: \textit{social media companies}} 
        \end{aligned} 
    \right\} \\
    &\land \, \text{Data subjects: \textit{social media users}} \\
    &\land \, \mathbf{\bigvee} \left\{ 
        \text{Application means: All except \textit{supplant human moderators}, \textit{automated removal}}
    \right\} \\
    &\land \, \mathbf{\bigvee} \left\{ 
        \text{Model means: All except \textit{human in the loop}} 
    \right\}
\end{align*}

\vspace{0.5em}

\noindent\textbf{Vague Opposition:}
\begin{align*}
    &\text{Application means: \textit{not specified}} \\
    &\land \, \text{Ends: \textit{fight hate}}
\end{align*}
\subsection{Logical rule derivation example}
\begin{figure}[H]
    \centering
    \includegraphics[width=.9\linewidth]{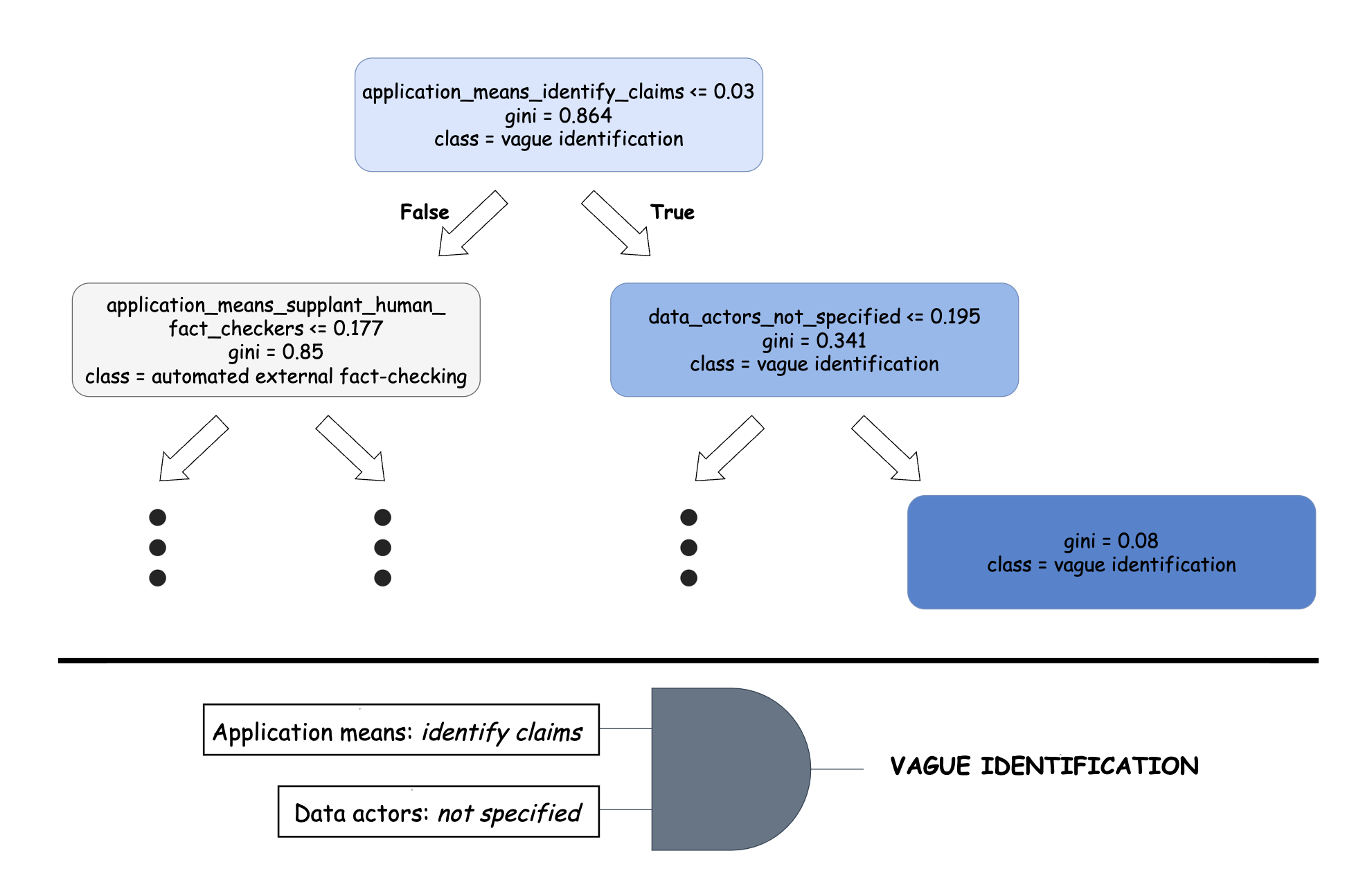}
    \caption{Example illustrating how the logical rule for \textit{Vague Identification} is derived from the decision tree.}
    \label{fig:tree-to-rule}
\end{figure}
\newpage
\twocolumn
\section{Research Framings Distribution in Annotations}
\subsection*{Hate Speech Dataset}
\label{app:hs-dist}
\begin{figure}[H]
    \centering
    \includegraphics[width=.9\linewidth]{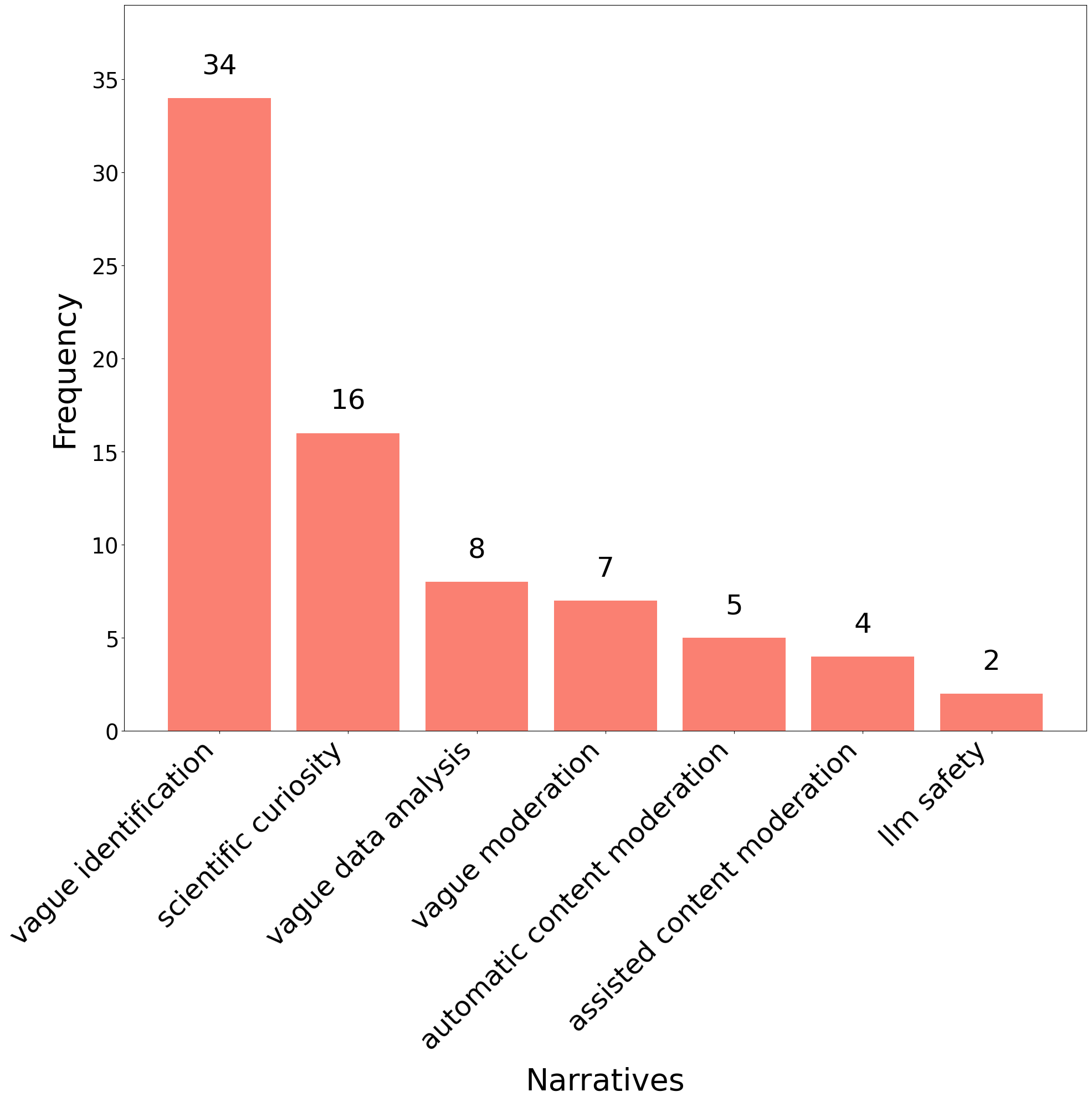}
    \caption{Distribution of research framings identified by human annotators in our collected hate speech detection papers.}
    \label{fig:labels-hsl}
\end{figure}
\subsection*{AFC Literature Analysis}
\label{app:afc-lit-dist}
\begin{figure}[H]
    \centering
    \includegraphics[width=\linewidth]{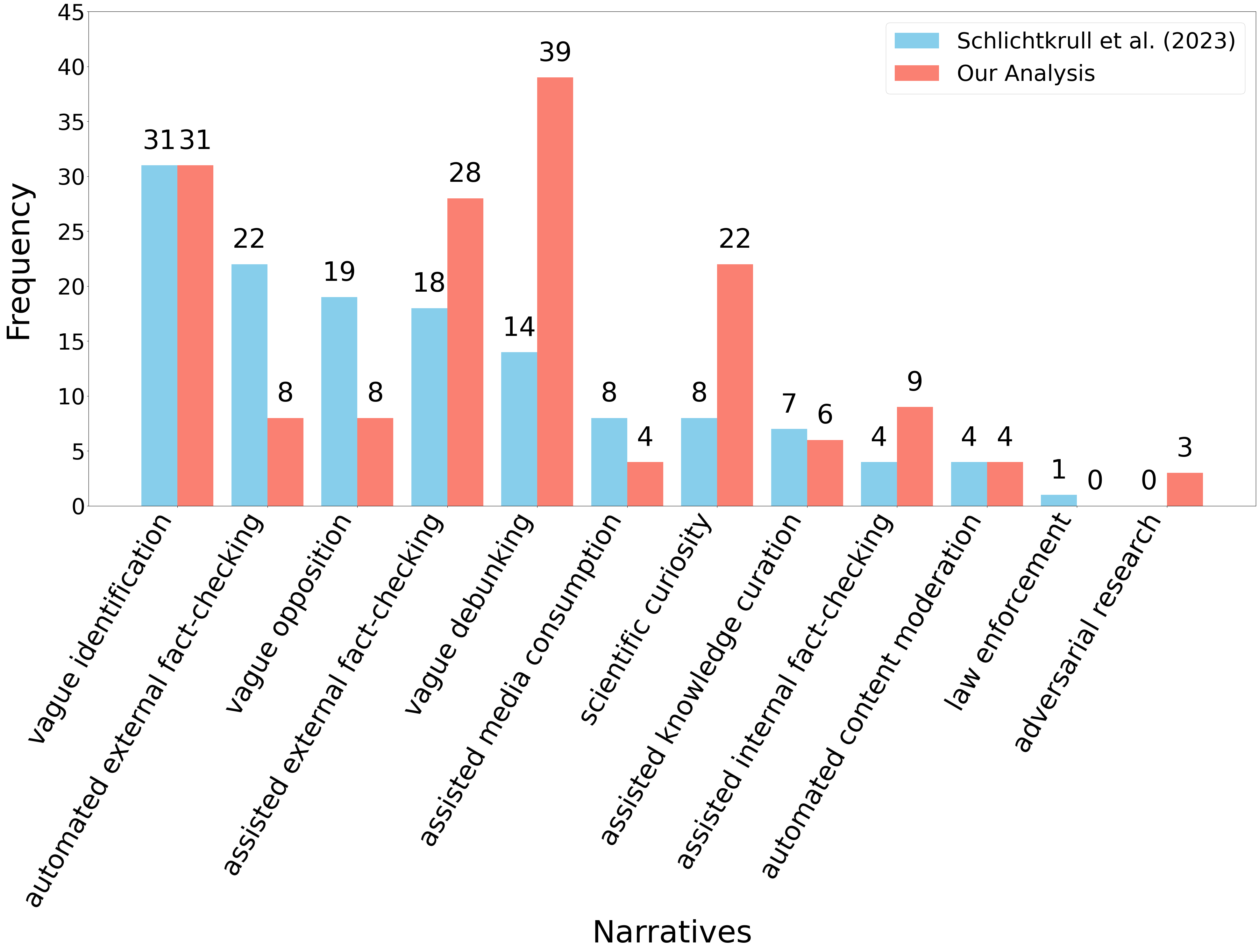}
    \caption{Comparison of the distribution of research framings in our analysis (2023 onwards) and that of \citet{intendeduses} (until 2023).}
    \label{fig:narrative-comp}
\end{figure}

\section{Hate Speech Annotation Guidelines}
\label{annotation-appendix}
To investigate the epistemology of hate speech detection and related tasks (e.g. bias in hate speech), we annotate the research framings of 49 research papers.

We adopt a two-step annotation process: (1) \textit{paragraph-level annotation} to identify excerpts related to the paper’s goals and methods, labeling \textit{data subjects}, \textit{data actors}, \textit{model means}, \textit{application means}, and \textit{epistemic ends}; and (2) \textit{discourse-level annotation} to infer the underlying research framings.

Our goal is to examine whether the means proposed in these research framings can realistically achieve the intended aims.

\subsection{Paper Selection}
We collect 49 research papers focused on hate speech detection and related tasks (e.g., bias in hate speech). Annotations are extracted from introductions and abstracts.

Initially, we annotated a small subset of papers based on predefined criteria. We refined category definitions through discussion, resolved label inconsistencies, and conducted a final annotation round to ensure consistency.

\subsection{Annotation Criteria}
We analyze each paper’s introduction and abstract, structuring epistemic research framings at two levels: (1) \textit{paragraph-level categorization} and (2) \textit{research framing extraction}.

\subsubsection{Paragraph-Level Annotation} 
At this level, we extract modeling and application means, epistemic goals, data actors, and data subjects. These elements are then combined into discourse-level research framings, e.g.:

\textit{``Social media moderators (actor) should use a classification model (modeling means) to triage posts (application means) from social media users (data subjects) to moderate online content (goal).''}

Our annotation process follows an inductive approach: we begin with a predefined set of categories based on an initial pilot study. If, during annotation, we encounter means, ends, actors, or research framings that do not fit existing categories, we introduce new ones. We use a multi-label approach, as paragraphs and discourses may reference multiple categories.

\subsubsection{Quotes} 
We extract epistemic quotes relevant to the paper’s research framings, focusing on the \textit{why} and \textit{what}. Quotes are primarily sourced from the introduction; missing information is supplemented from the abstract.

\subsubsection{Data Subjects} 
Entities whose texts are analyzed or moderated:

\begin{itemize}
    \item Social media users
    \item Journalists
    \item Public figures/politicians
    \item Product reviewers
    \item Media consumers
\end{itemize}

\paragraph{Social Media Users} Includes anonymous contributors to social media, excluding explicitly public figures.

\paragraph{Journalists} Covers professional journalists and online publishers but excludes amateur journalists.

\paragraph{Public Figures/Politicians} Includes public figures such as politicians, actors, or influencers.

\paragraph{Product Reviewers} Individuals posting reviews on platforms like Amazon or Trustpilot.

\paragraph{Media consumers} End-users consuming content verified or enhanced by model outputs.

\subsubsection{Data Actors} 
Entities intended to act on model outputs:

\begin{itemize}
    \item Social media moderators
    \item Social media companies
    \item Scientists
    \item Media consumers
    \item Engineers and curators
    \item Law enforcement
    \item Algorithm
\end{itemize}

\paragraph{Social Media Moderators} Individuals employed to moderate social media content. Applicable only when explicitly stated that humans will act on model outputs.

\paragraph{Social Media Companies} Covers broader company-level decisions, including automated interventions based on model outputs.

\paragraph{Scientists} Includes researchers who analyze model outputs for scientific research rather than direct intervention.

\paragraph{Media Consumers} General users who encounter model outputs in content filtering systems.

\paragraph{Engineers and Curators} Covers those maintaining curated knowledge repositories such as Wikipedia.

\paragraph{Law Enforcement} Includes police, intelligence agents, or legal professionals utilizing model outputs.

\paragraph{Algorithm} Refers to fully automated systems that act on model outputs without human oversight.

\subsubsection{Model Owners} 
Entities responsible for owning or deploying the models:

\begin{itemize}
    \item Media companies
    \item Social media companies
    \item Law enforcement
\end{itemize}

\paragraph{Media Companies} Includes journalists and fact-checkers working for professional organizations.

\paragraph{Social Media Companies} Covers engineers and executives responsible for deploying moderation systems.

\paragraph{Law Enforcement} Includes agencies using model outputs for investigative or judicial purposes.

\subsubsection{Modeling (ML) Means} 
The technical approach proposed in the paper:

\begin{itemize}
    \item Classify/score posts or text
    \item Provide justifications
    \item Corpora analysis
    \item Data collection
    \item Human-in-the-loop
\end{itemize}

\paragraph{Classify/Score Posts or Text} 
When the paper proposes classifying posts (e.g., hate vs. non-hate, toxic vs. non-toxic) or using classifier-generated scores.

\paragraph{Provide Justifications} 
When the paper proposes generating explanations or justifications to support its goal (\textit{epistemic ends}).

\paragraph{Corpora Analysis} 
When the methodology involves analyzing datasets or performing large-scale linguistic analysis.

\paragraph{Data Collection} 
When the paper focuses on collecting new datasets to address a specific problem.

\paragraph{Human-in-the-Loop} 
When human actors are integrated into the model’s workflow, either as decision-makers or contributors.

\subsubsection{Application Means} 
How the model is applied:

\begin{itemize}
    \item Identify toxic content
    \item Supplant human moderators
    \item Gather and present justifications
    \item Automated removal
    \item Provide labels or toxicity scores
    \item Filter system outputs
    \item Analyze data
    \item Vague persuasion
    \item Generate counter narratives
\end{itemize}

\paragraph{Identify Toxic Content} 
ML models should identify harmful or offensive content from large volumes of online posts.

\paragraph{Supplant Human Moderators} 
Fully or partially replace human moderators by automating content moderation. If a human-in-the-loop approach is intended, this label is not applied.

\paragraph{Gather and Present Justifications} 
ML models should generate and present relevant justifications for why a post is harmful.

\paragraph{Automated Removal} 
ML models should autonomously remove posts from social media or other platforms without human intervention.

\paragraph{Provide Labels or Toxicity Scores} 
ML models should assign toxicity labels or scores to content. The target audience could be either users or platform administrators.

\paragraph{Filter System Outputs} 
ML models should act as filters to ensure that outputs from other models (e.g., LLMs) are free from toxicity or misinformation.

\paragraph{Analyze Data} 
Models and datasets should be used for research purposes, such as understanding the spread of hate speech or toxicity.

\paragraph{Vague Persuasion} 
ML models should influence users' perspectives through mechanisms such as warning labels or counter-narratives, though the exact approach is unspecified.

\paragraph{Generate counter narratives}: If the excerpt explicitly mentions that the released artefacts are designed to produce responses or statements aimed at refuting, mitigating, or opposing harmful or hateful content (e.g., generating automated replies to hate speech).

\subsubsection{Epistemic Ends} 
The overarching goal of the research:

\begin{itemize}
    \item Fight hate
    \item Counter hate
    \item Limit AI-generated toxicity
    \item Develop knowledge of NLP/language
    \item Avoid biases of human moderators
    \item Detect hate for law enforcement
\end{itemize}

\paragraph{Fight Hate} 
The ultimate aim is to reduce the spread or influence of hate speech.

\paragraph{Counter Hate} 
The goal is to challenge stereotypes and harmful narratives.

\paragraph{Limit AI-Generated Toxicity} 
The objective is to mitigate the risks of AI-generated harmful or toxic content.

\paragraph{Develop Knowledge of NLP/Language} 
Research is conducted to better understand language structures and improve NLP models.

\paragraph{Avoid Biases of Human Moderators} 
The aim is to develop moderation systems that overcome human biases.

\paragraph{Detect Hate for Law Enforcement} 
The model is intended for use in legal or law enforcement contexts, such as aiding police investigations or courtroom decisions.

\subsubsection{Important Note} 
We rely strictly on explicit statements in extracted quotes, avoiding inferred or implied meanings.

\subsection{Research Framings}
\label{new-narratives-appendix}
     At the discourse level, we extract the epistemic research framings present in the paper. We envision research framings as discourse structures combining the (modeling/application) means, ends, data actors, and data subjects extracted at the paragraph level.

\begin{table}[]
    \centering
    \begin{tabular}{c|c}
    \toprule
       \textbf{Research Framings} & \textbf{Ordinal Class} \\
       \midrule
       Vague Opposition  & -3  \\
       Vague Identification  & -2  \\
       Vague Moderation  & -1  \\
       Law Enforcement & 0 \\
       Scientific Curiosity  & 1  \\
       Automatic Content Moderation  & 2  \\
       Assisted Content Moderation  & 3  \\
       Assisted Knowledge Curation & 4 \\
    \bottomrule
    \end{tabular}
    \caption{Ordinal classes for research framings: underspecified research framings have negative values.}
    \label{tab:ordinal_classes}
\end{table}

\paragraph*{LLM Safety} 
The paper mentions limiting toxic publishes or generated content as an end.
\paragraph*{Vague Identification} 
The paper mentions identifying or detecting toxic posts as the means, and fighting hate as the end. However, it is not clear how the authors intend to accomplish that end using proposed means. Typically, there are no data actors -- it is also not clear who should act on the model predictions.

\paragraph*{Vague Moderation} 
When the paper proposes to assist the moderation process without explaining how. It is clear that the mechanism is supposed to follow what moderators are currently doing, but not where or how the ML model will be used in this process. Furthermore, it is \textit{unclear} whether the entire process will be automated. For example, the paper may suggest that a model should be used by moderators, but it is not clear \textit{how} the model assists in that. A common warning sign is a mismatch between ML means and application means: classification is not really useful for moderators (or social media users), and so papers suggesting to use those either to assist or automate moderation are often not clear in how.

\paragraph*{Vague Opposition} 
Restricted to cases without any application means (model means and no application means in contrast to vague identification and vague debunking). E.g. \textit{a machine learning/automated system will reduce the spread of misinformation.}

When the paper presents a research framing of vague opposition to misinformation. The \textit{end} is to limit the spread or influence of misinformation, and the \textit{ML means} are, for example, to classify claims. However, the connection between means and ends is left unmentioned, and epistemic actors are typically entirely absent. An impression is given that the development of automated fact-checking will limit the spread of misinformation, but the link between the two is left unstated. 

\paragraph*{Assisted Content Moderation} 
If the paper proposes the deployment of automated moderation as a tool to assist content moderators on social media platforms. Here, the \textit{end} is to fight hate on social media platforms, the \textit{means} are to provide suggestions for posts to delete (along with, potential justifications/explanations), and the \textit{actors} are human moderators who make the final choice on whether posts should be deleted or not.

\paragraph*{Automatic Content Moderation} 
If the paper analysed proposes a similar content moderation strategy, but instead of assisting human moderator, it suggests replacing them entirely. In this case, the \textit{end} is to fight hate online, the \textit{means} are to deploy classifiers to flag posts and remove any labelled offensive, and the \textit{actors} are the executives and engineers at social media companies who deploy and make decisions about such systems.

\paragraph*{Assisted Knowledge Curation} 
When the analysed paper proposes toxic content detection primarily as a component filtering the information kept in some curated knowledge vault, including graph-based knowledge bases as well as text-based collections such as Wikipedia. Here, the \textit{end} is to avoid toxicity/stereotypical content of the knowledge vault, the \textit{means} are to use automated fact-checking as an additional toxic content filter layer that prevents disputed content from being added (or interrogates already added content), and the \textit{actors} are typically the engineers who maintain the knowledge base.

\paragraph*{Scientific Curiosity} 
When the authors of the analysed paper justified their projects purely on the basis of scientific curiosity. While differing strongly from the other research framings presented here, this is still a virtue epistemological research framing, concerned with the production of \textit{good knowledge}. Here, the \textit{end} is to increase scientific knowledge of semantics; the \textit{means} are to learn how to build automated systems that mimic human moderators, a process theorised to yield knowledge about the construction of meaning; and the \textit{actors} are scientists in natural language processing and adjacent fields.

\paragraph*{Vague Data Analysis} 
When the authors of the analysed paper justify their data analysis on the basis of scientific curiosity but also claim that the end goal is to fight hate without a clear explanation about how to achieve this.

\section{Prompts}
\label{prompts-appendix}
\subsection{Epistemic Element Extraction: Hate speech detection}
You will be provided with the introduction of a computer science research paper on hate speech detection and related tasks to understand the broader context of the study. Afterward, you will be given a specific excerpt from the paper. Your task is to answer the question based solely on the specific excerpt while considering the broader context provided in the introduction.

\textbf{Introduction:} \texttt{\{introduction\}}

\textbf{Specific Excerpt:} \texttt{\{excerpt\}}

\subsubsection{Data Subjects}
\begin{quote}
\noindent
\textbf{Task:} Identify the \textbf{data subjects} explicitly mentioned by the authors in the provided excerpt. Data subjects are individuals or entities that the authors explicitly identify as the intended targets of the artefacts released in this study. Your answer should be based solely on the provided excerpt.

\textbf{Key Rules:}
\begin{enumerate}
    \item \textbf{Explicit mention only}: Consider only data subjects explicitly named or described in the excerpt.
    \item \textbf{Focus on the released artefacts}: Differentiate between general background information and what the authors specifically propose.
    \item \textbf{Default to 'not specified in this paragraph'}: If no data subjects are explicitly mentioned, respond accordingly.
    \item \textbf{Consider the big picture}: Identify who the authors envision as targets for their released artefacts.
\end{enumerate}

\textbf{Question:} Based on the provided excerpt and using the provided labels, who are the explicitly mentioned data subjects?

You will select the single most appropriate label that answers the question. Provide reasoning then choose only from the following list:

\begin{itemize}
    \item \textbf{Journalists}: If the excerpt explicitly mentions that the released artefacts are designed to be applied on statements from journalists, including online publishers.
    \item \textbf{Social media users}: If the excerpt explicitly mentions that the released artefacts are designed to be applied on statements from contributors to social media platforms, excluding public figures. This includes commenters on forums as well as, e.g., Twitter users. Similarly, editors of Wikipedia and other collaborative writing projects fall under this category as well.
    \item \textbf{Public figures/politicians}: If the excerpt explicitly mentions that the released artefacts are designed to be applied on statements from public figures like politicians or actors. These statements could be from media releases, interviews, speeches, or similar.
    \item \textbf{Product reviewers}: If the excerpt explicitly mentions that the released artefacts are designed to be applied on statements from individuals reviewing products on platforms like Amazon or Trustpilot.
    \item \textbf{Media consumers}: If the excerpt explicitly mentions that the released artefacts are expected to be applied on statements by end-users consuming content verified or enhanced by model outputs.
\end{itemize}

If none of these labels apply, respond with "not specified in this paragraph."
\end{quote}

\subsubsection{Data Actors}
\begin{quote}
\textbf{Task}:
Identify the \textbf{data actors} explicitly mentioned in the provided excerpt. Data actors are individuals or entities expected to act on or utilize the artefacts released in this study.

\textbf{Key Rules:}
\begin{enumerate}
    \item \textbf{Explicit mention only}: Consider only explicitly named data actors.
    \item \textbf{Focus on the released artefacts}: Ensure relevance to the specific artefacts described in the excerpt.
    \item \textbf{Default to 'not specified in this paragraph'}: If no data actors are explicitly mentioned, respond accordingly.
    \item \textbf{Consider the big picture}: Identify who the authors expect to act on the model outputs.
\end{enumerate}

\textbf{Question:} Based on the provided excerpt and using the provided labels, who are the explicitly mentioned data actors?
\end{quote}
\begin{quote}
\noindent
Identify the individuals or entities expected to act on the model outputs in the excerpt.

\textbf{Possible Labels:}
\begin{itemize}
    \item \textbf{Social media moderators}: If the excerpt explicitly mentions that the released artefacts are expected to be utilized by individuals hired to moderate social media spaces, i.e., explicitly stated that a human employee should act on model outputs.
    \item \textbf{Social media companies}: If the excerpt explicitly mentions that the released artefacts are expected to be utilized by social media companies, including engineers working to maintain the network. This is applicable when decisions should be made based on model outputs or when it is unclear.
    \item \textbf{Scientists}: If the excerpt explicitly mentions that the released artefacts are expected to be utilized by researchers using model outputs for scientific analysis or experiments. E.g., to analyze data with the express purpose of learning something about it, not acting on the model decisions.
    \item \textbf{Media consumers}: If the excerpt explicitly mentions that the released artefacts are expected to be utilized by end-users consuming content verified or enhanced by model outputs. Only applicable when the consumer is directly understood (possibly implicitly) to use the system, e.g., in the form of a browser extension. The decision to use the outputs of the tool must be in the hands of the consumer.
    \item \textbf{Engineers and curators}: If the excerpt explicitly mentions that the released artefacts are expected to be utilized by maintainers of knowledge bases integrating model outputs.
    \item \textbf{Law enforcement}: If the excerpt explicitly mentions that the released artefacts are expected to be utilized by agents using model outputs in investigations or legal processes.
    \item \textbf{Algorithm}: If the excerpt explicitly mentions that the released artefacts are expected to be utilized by automated systems acting directly on model outputs without human intervention. E.g., remove posts based on the model's predictions.
\end{itemize}
If none of these labels apply, respond with "not specified in this paragraph."
\end{quote}

\subsubsection{Model Owners}
\begin{quote}
\textbf{Task}:
Identify the \textbf{model owners} explicitly mentioned by the authors in the provided excerpt. Model owners are entities responsible for controlling or maintaining the artefacts released in this study.

\textbf{Key Rules:}
\begin{enumerate}
    \item \textbf{Explicit mention only}: Consider only model owners explicitly named in the excerpt.
    \item \textbf{Focus on the released artefacts}: Do not assume ownership unless explicitly stated.
    \item \textbf{Default to 'not specified in this paragraph'}: If no entities are explicitly mentioned as owners, respond accordingly.
    \item \textbf{Consider the big picture}: Identify the responsible institutions or organizations.
\end{enumerate}

\textbf{Question:} Based on the provided excerpt and using the provided labels, who are the explicitly mentioned model owners?

\noindent
Determine the entities responsible for maintaining the model.

\textbf{Possible Labels:}
\begin{itemize}
    \item \textbf{Media companies}: If the excerpt explicitly mentions that the released artefacts are expected to be owned by journalists and hate speech detection entities working for professional (non-amateur) media companies.
    \item \textbf{Social media companies}: If the excerpt explicitly mentions that the released artefacts are expected to be owned by organizations maintaining social platforms and using models for moderation, including the engineers working to maintain the social network.
    \item \textbf{Law enforcement}: If the excerpt explicitly mentions that the released artefacts are expected to be owned by agencies employing models for investigations and truth verification.
\end{itemize}
If none of these labels apply, respond with "not specified in this paragraph."
\end{quote}

\subsubsection{Modeling (ML) Means}
\begin{quote}
\textbf{Task}:
Identify the \textbf{modeling means} explicitly mentioned in the provided excerpt. Modeling means refer to the specific machine learning methods or strategies proposed by the authors.

\textbf{Key Rules:}
\begin{enumerate}
    \item \textbf{Explicit mention only}: Consider only modeling methods explicitly named in the excerpt.
    \item \textbf{Focus on the released artefacts}: Do not infer methods unless explicitly stated.
    \item \textbf{Default to 'not specified in this paragraph'}: If no methods are explicitly mentioned, respond accordingly.
    \item \textbf{Consider the big picture}: Identify the modeling strategies used in the study.
\end{enumerate}

\textbf{Question:} Based on the provided excerpt and using the provided labels, what are the explicitly mentioned modeling means?

\noindent
Identify the machine learning methods proposed in the excerpt.

\textbf{Possible Labels:}
\begin{itemize}
    \item \textbf{Classify/score posts/text}: If the excerpt explicitly mentions that the released artefacts involve classifying posts (hate/toxic and non-hate/non-toxic etc.) or using the scores returned by the classifier.
    \item \textbf{Provide justifications}: If the excerpt explicitly mentions that released artefacts involve generating explanations for model decisions to reach its goals (ends).
    \item \textbf{Human in the loop}: If the excerpt explicitly mentions that released artefacts involve including human oversight in the solution or main process described in the paper.
    \item \textbf{Corpora analysis}: If the excerpt explicitly mentions that released artefacts involve using data analytics or analyzing datasets for patterns or insights.
    \item \textbf{Data collection}: If the excerpt explicitly mentions that released artefacts involve collecting data to solve a given problem.
\end{itemize}
If none of these labels apply, respond with "not specified in this paragraph."
\end{quote}

\subsubsection{Application Means}
\begin{quote}
\textbf{Task}:
Identify the \textbf{application means} explicitly mentioned by the authors in the provided excerpt. Application means refer to the concrete, actionable uses of the artefacts released in the study.

\textbf{Key Rules:}
\begin{enumerate}
    \item \textbf{Explicit mention only}: Consider only application means explicitly described in the excerpt.
    \item \textbf{Focus on the released artefacts}: Differentiate between general discussion and specific proposals.
    \item \textbf{Default to 'not specified in this paragraph'}: If no application means are mentioned, respond accordingly.
    \item \textbf{Consider the big picture}: Identify how the authors envision their artefacts being applied.
\end{enumerate}

\textbf{Question:} Based on the provided excerpt and using the provided labels, what are the explicitly mentioned application means?

\noindent
Identify the concrete application of the model.

\textbf{Possible Labels:}
\begin{itemize}
    \item \textbf{Identify toxic content}: If the excerpt explicitly mentions that released artefacts would be leveraged to detect potentially harmful/offensive posts on the Internet.
    \item \textbf{Supplant human moderators}: If the excerpt explicitly mentions that released artefacts would be leveraged to replace human moderators entirely or at least partially by automatically and independently handling some content in real-world scenarios. If the intention is a human-in-the-loop system, supplanting human moderators is not the means.
    \item \textbf{Gather and present justification}: If the excerpt explicitly mentions that released artefacts would be leveraged to find relevant justification for why a post is harmful, with the aim of showing it to a human.
    \item \textbf{Automated removal}: If the excerpt explicitly mentions that released artefacts would be leveraged to automatically delete or hide claims identified as false, harmful, or in violation of platform policies.
    \item \textbf{Provide labels/toxicity scores}: If the excerpt explicitly mentions that released artefacts would be leveraged to assign toxicity scores, labels, or indicators to individual claims and show them to users/human annotators.
    \item \textbf{Filter system outputs}: If the excerpt explicitly mentions that released artefacts would be leveraged to refine or validate the outputs of other machine learning systems to improve accuracy or alignment.
    \item \textbf{Analyze data}: If the excerpt explicitly mentions that released artefacts would be leveraged to examine and study patterns or trends in datasets of content related to toxicity or hate speech detection to get a better understanding of how hate/toxicity spreads.
    \item \textbf{Vague persuasion}: If the excerpt explicitly mentions that released artefacts would be leveraged to influence user decisions or beliefs (e.g., through warning labels) without specifying the exact method of persuasion.
    \item \textbf{Generate counter-narratives}: If the excerpt explicitly mentions that the released artefacts are designed to produce responses or statements aimed at refuting, mitigating, or opposing harmful or hateful content (e.g., generating automated replies to hate speech).
\end{itemize}
If none of these labels apply, respond with "not specified in this paragraph."
\end{quote}
\newpage
\subsubsection{Ends}
\textbf{Task}:
\begin{quote}
Identify the \textbf{ends} explicitly mentioned by the authors in the provided excerpt. The ends represent the ultimate goals or outcomes of the released artefacts.

\textbf{Key Rules:}
\begin{enumerate}
    \item \textbf{Explicit mention only}: Consider only explicitly mentioned goals or outcomes.
    \item \textbf{Focus on the released artefacts}: Do not assume end goals unless explicitly stated.
    \item \textbf{Default to 'not specified in this paragraph'}: If no end goals are mentioned, respond accordingly.
    \item \textbf{Consider the big picture}: Identify the broader societal or epistemic impact.
\end{enumerate}

\textbf{Question:} Based on the provided excerpt and using the provided labels, what are the explicitly mentioned ends?
\noindent
Determine the ultimate goal of the study.

\textbf{Possible Labels:}
\begin{itemize}
    \item \textbf{Fight hate}: If the excerpt explicitly mentions that the ultimate goal of the released artefacts would be to prevent hate from spreading or to limit its influence.
    \item \textbf{Counter hate}: If the excerpt explicitly mentions that the ultimate goal of the released artefacts would be to counter stereotypes and harmful narratives.
    \item \textbf{Limit AI-generated toxicity/misinformation}: If the excerpt explicitly mentions that the ultimate goal of the released artefacts would be to prevent AI-generated harmful/toxic content from spreading.
    \item \textbf{Develop knowledge of NLP/language}: If the excerpt explicitly mentions that the ultimate goal of the released artefacts would be to advance understanding of language and NLP methods. Authors would be studying the difficult problem of hate speech detection to learn more about how language works and how to build models that interpret semantics.
    \item \textbf{Avoid biases of human moderators}: If the excerpt explicitly mentions that the ultimate goal of the released artefacts would be to achieve unbiased and consistent hate speech detection.
    \item \textbf{Detect hate/falsehood for law enforcement}: If the excerpt explicitly mentions that the ultimate goal of the released artefacts would be to enable law enforcement to identify and act on hate speech.
\end{itemize}
If none of these labels apply, respond with "not specified in this paragraph."
\end{quote}
\subsection{Epistemic Element Extraction: Automated Fact-checking}
You will be provided with the introduction of a computer science research paper on automated fact-checking to understand the broader context of the study. 
Afterward, you will be given a specific excerpt from the paper. Your task is to answer the question based solely on the specific excerpt while considering the broader context provided in the introduction.

\begin{quote}
\textbf{Introduction:} \texttt{\{introduction\}} 

\textbf{Specific Excerpt:} \texttt{\{excerpt\}}

\subsubsection{Data Subjects}

\noindent
\textbf{Task:} 
Identify the \textbf{data subjects} explicitly mentioned by the authors in the provided excerpt. Data subjects are the individuals or entities that the authors explicitly identify as the intended targets of the artefacts released in this study. The possible data subjects will be outlined below, and your answer should solely be based on the provided excerpt.

\paragraph{Key rules:}
\begin{enumerate}
    \item \textbf{Explicit mention only}: Consider only data subjects explicitly named or described in the excerpt. Do not infer their presence unless they are directly identified as targeted by the released artefacts.
    \item \textbf{Focus on the released artefacts}: Differentiate between general background information and what the authors specifically propose. If the authors discuss fact-checking broadly or reference prior work, do not assume relevance; focus solely on the data subjects tied to the artefacts they release. For example, if the text discusses fake news or its impact, do not assume social media users are data subjects unless explicitly mentioned as being the intended targets. Similarly, if politics is mentioned, do not assume politicians are data subjects unless they are explicitly identified.
    \item \textbf{Default to 'not specified in this paragraph'}: If no data subjects are explicitly mentioned or described as targeted to use the released artefacts, respond with \textbf{"not specified in this paragraph"}.
    \item \textbf{Consider the big picture}: The goal is to automatically identify who the authors envision as targets for their released automated fact-checking artefacts, if mentioned at all. Consider the study's broader purpose and how it is represented in the specific excerpt.
\end{enumerate}

\textbf{Question:} Based on the provided excerpt and using the provided labels, who are the explicitly mentioned data subjects?

\end{quote}

You will select the single most appropriate label that answers the question. 
Provide reasoning then choose only from the following list:

\begin{itemize}
    \item \textbf{Professional Journalists}: If the excerpt explicitly mentions that the released artefacts are designed to be applied on statements from professional journalists, including fact-checkers.
    \item \textbf{Citizen Journalists}: If the excerpt explicitly mentions that the released artefacts are designed to be applied on statements from amateur journalists or collectives working without formal training.
    \item \textbf{Social Media Users}: If the excerpt explicitly mentions that the released artefacts are designed to be applied on statements from contributors to social media platforms, excluding public figures.
    \item \textbf{Technical Writers}: If the excerpt explicitly mentions that the released artefacts are designed to be applied on statements from writers of technical or scientific documents, including legal professionals, to spot errors in their articles before publication.
    \item \textbf{Public Figures/Politicians}: If the excerpt explicitly mentions that the released artefacts are designed to be applied on statements from public figures like politicians or actors.
    \item \textbf{Product Reviewers}: If the excerpt explicitly mentions that the released artefacts are designed to be applied on statements from individuals reviewing products on platforms like Amazon or Trustpilot.
    \item \textbf{Suspects}: If the excerpt explicitly mentions that the released artefacts are designed to be applied on statements from people under investigation or scrutiny, often in a legal or forensic context.
\end{itemize}

If none of these labels apply, respond with "not specified in this paragraph."
\newpage
\subsubsection{Data Actors}
\noindent
\textbf{Task:} Identify the \textbf{data actors} explicitly mentioned in the provided excerpt. Data actors are individuals or entities that the authors specifically identify as expected to act on or utilize the artefacts released in this study. The possible data actors will be outlined below, and your answer should solely be based on the provided excerpt.

\paragraph{Key rules:}
\begin{enumerate}
    \item \textbf{Explicit mention only}: Consider only data actors explicitly named or described in the excerpt. Do not infer their presence unless they are directly identified as acting on the released artefacts.
    \item \textbf{Focus on the released artefacts}: Differentiate between general background information and what the authors specifically propose. If the authors discuss fact-checking broadly or reference prior work, do not assume relevance; focus solely on the data actors tied to the artefacts they release. For example, if the text discusses fake news or its impact, do not assume journalists, moderators, or consumers are data actors unless explicitly mentioned as acting on the model's outputs. Similarly, if a suspect or individual is mentioned, do not assume they are a data actor unless explicitly described as targeted to interact with the model.
    \item \textbf{Default to 'not specified in this paragraph'}: If no data actors are explicitly mentioned or described as targeted to use the released artefacts, respond with \textbf{"not specified in this paragraph"}.
    \item \textbf{Consider the big picture}: The goal is to automatically identify who the authors envision as actors for their released automated fact-checking artefacts, if mentioned at all.
\end{enumerate}

\textbf{Question:} Based on the provided excerpt and using the provided labels, who are the explicitly mentioned data actors?

You will select the single most appropriate label that answers the question. 
Provide reasoning then choose only from the following list:
\noindent

\begin{itemize}
    \item \textbf{Publishers}: If the excerpt explicitly mentions that the released artefacts are expected to be utilized by organizations or individuals that distribute printed or online content, such as books and comics.
    \item \textbf{Professional journalists}: If the excerpt explicitly mentions that the released artefacts are expected to be utilized by professional journalists, including fact-checkers and reporters to get model insights on their work.
    \item \textbf{Product reviewers}: If the excerpt explicitly mentions that the released artefacts are expected to be utilized by reviewers providing insights or opinions about products.
    \item \textbf{Citizen journalists}: If the excerpt explicitly mentions that the released artefacts are expected to be utilized by amateur reporters leveraging model outputs for investigative purposes.
    \item \textbf{Scientists}: If the excerpt explicitly mentions that the released artefacts are expected to be utilized by researchers using model outputs for scientific analysis or experiments.
    \item \textbf{Media consumers}: If the excerpt explicitly mentions that the released artefacts are expected to be utilized by end-users consuming content verified or enhanced by model outputs.
    \item \textbf{Technical writers}: If the excerpt explicitly mentions that the released artefacts are expected to be utilized by writers improving technical or legal documents with model suggestions.
    \item \textbf{Engineers and curators}: If the excerpt explicitly mentions that the released artefacts are expected to be utilized by maintainers of knowledge bases integrating model outputs.
    \item \textbf{Law enforcement}: If the excerpt explicitly mentions that the released artefacts are expected to be utilized by agents using model outputs in investigations or legal processes.
    \item \textbf{Algorithm}: If the excerpt explicitly mentions that the released artefacts are expected to be utilized by automated systems acting directly on model outputs without human intervention.
\end{itemize}

If none of these labels apply, respond with "not specified in this paragraph."

\subsubsection{Model Owners}
\textbf{Task:} Identify the \textbf{model owners} explicitly mentioned by the authors in the provided excerpt. Model owners are entities or organisations that the authors explicitly target as being responsible for controlling or maintaining the artefacts released in this study. The possible model owners will be outlined below, and your answer should solely be based on the provided excerpt.

\paragraph{Key rules:}
\begin{enumerate}
    \item \textbf{Explicit mention only}: Consider only model owners explicitly named or described in the excerpt. Do not infer their presence unless they are directly identified as expected to own the released artefacts.
    \item \textbf{Focus on the released artefacts}: Differentiate between general background information and what the authors specifically propose. If the authors discuss fact-checking broadly or reference prior work, do not assume relevance; focus solely on the model owners tied to the artefacts they release. For example, if the text mentions researchers conducting experiments, it does not imply they own the models unless explicitly stated.
    \item \textbf{Default to 'not specified in this paragraph'}: If no entities or organisations are explicitly mentioned or intended to own the released artefacts, respond with \textbf{"not specified in this paragraph"}.
    \item \textbf{Consider the big picture}: The goal is to automatically identify who the authors envision as owning their released automated fact-checking artefacts, if mentioned at all.
\end{enumerate}

\textbf{Question:} Based on the provided excerpt and using the provided labels, who are the explicitly mentioned model owners?

\noindent
The task follows the same structure as for Data Subjects, but with the following labels:

\begin{itemize}
    \item \textbf{Social media companies}: If the excerpt explicitly mentions that the released artefacts are expected to be owned by organizations maintaining social platforms and using models for moderation.
    \item \textbf{Law enforcement}: If the excerpt explicitly mentions that the released artefacts are expected to be owned by agencies employing models for investigations and truth verification.
    \item \textbf{Scientists}: If the excerpt explicitly mentions that the released artefacts are expected to be owned by researchers developing or owning models for analysis and experiments.
\end{itemize}

If none of these labels apply, respond with \textbf{"not specified in this paragraph"}.

\subsubsection{Model Means}
\textbf{Task:} Identify the \textbf{model means} explicitly mentioned by the authors in the provided excerpt. Model means are the specific machine learning methods or strategies proposed by the authors in their study. The possible model means will be outlined below, and your answer should solely be based on the provided excerpt.

\paragraph{Key rules:}
\begin{enumerate}
    \item \textbf{Explicit mention only}: Consider only model means explicitly named or described in the excerpt. Do not infer their presence unless they are directly identified.
    \item \textbf{Focus on the released artefacts}: Differentiate between general background information and what the authors specifically propose. If the authors discuss fact-checking broadly or reference prior work, do not assume relevance; focus solely on the model means tied to the artefacts they release.
    \item \textbf{Default to 'not specified in this paragraph'}: If no specific methods or strategies are mentioned in the excerpt, respond with \textbf{"not specified in this paragraph"}.
    \item \textbf{Consider the big picture}: Identify the strategies used by the authors in their study, if mentioned at all.
\end{enumerate}

\textbf{Question:} Based on the provided excerpt and using the provided labels, who are the explicitly mentioned model means?

\noindent
You will select the single most appropriate label that answers the question. 
Provide reasoning then choose only from the following list:

\begin{itemize}
    \item \textbf{Classify/score veracity}: If the excerpt explicitly mentions that the released artefacts involve determining if claims are true or false.
    \item \textbf{Classify/score stance}: If the excerpt explicitly mentions that the released artefacts involve evaluating whether evidence supports or opposes claims.
    \item \textbf{Evidence retrieval}: If the excerpt explicitly mentions that released artefacts involve locating relevant evidence to support or refute claims.
    \item \textbf{Justification/explanation production}: If the excerpt explicitly mentions that released artefacts involve generating explanations for model decisions.
    \item \textbf{Corpora analysis}: If the excerpt explicitly mentions that released artefacts involve analyzing datasets for patterns or insights.
    \item \textbf{Human in the loop}: If the excerpt explicitly mentions that released artefacts involve including human oversight in the solution or main process described in the paper.
    \item \textbf{Generate claims}: If the excerpt explicitly mentions that released artefacts involve producing synthetic claims for adversarial training or testing.
    \item \textbf{Triage claims}: If the excerpt explicitly mentions that released artefacts involve ranking claims by importance for efficient processing.
\end{itemize}

If none of these labels apply, respond with "not specified in this paragraph."

\subsubsection{Application Means}
\textbf{Task:}Identify the \textbf{application means} explicitly mentioned by the authors in the provided excerpt. Application means refer to the concrete, actionable uses of the artefacts released in the study, as proposed by the authors, to directly support the study's ultimate goal. The possible application means will be outlined below, and your answer should solely be based on the provided excerpt.

\paragraph{Key rules:}
\begin{enumerate}
    \item \textbf{Explicit mention only}: Consider only application means explicitly named or described in the excerpt.
    \item \textbf{Focus on the released artefacts}: Differentiate between general background information and what the authors specifically propose.
    \item \textbf{Default to 'not specified in this paragraph'}: If no application means are mentioned in the excerpt, respond with \textbf{"not specified in this paragraph"}.
    \item \textbf{Consider the big picture}: Identify how the authors envision their automated fact-checking artefacts to be applied in a real-world scenario.
\end{enumerate}
\noindent
You will select the single most appropriate label that answers the question. 
Provide reasoning then choose only from the following list:

\begin{itemize}
    \item \textbf{Identify claims}: If the excerpt explicitly mentions that released artefacts would be leveraged to detect potentially false or misleading statements.
    \item \textbf{Triage claims}: If the excerpt explicitly mentions that released artefacts would be leveraged to prioritize claims based on urgency or relevance.
    \item \textbf{Supplant human fact-checkers}: If the excerpt explicitly mentions that released artefacts would be leveraged to replace human fact-checkers entirely or partially.
    \item \textbf{Gather and present evidence}: If the excerpt explicitly mentions that released artefacts would be leveraged to retrieve and organize evidence supporting or refuting specific claims.
    \item \textbf{Identify multimodal inconsistencies}: If the excerpt explicitly mentions that released artefacts would be leveraged to detect contradictions between different content types.
    \item \textbf{Automated removal}: If the excerpt explicitly mentions that released artefacts would be leveraged to automatically delete or hide claims identified as false or harmful.
    \item \textbf{Provide labels/veracity scores}: If the excerpt explicitly mentions that released artefacts would be leveraged to assign truthfulness scores or labels.
    \item \textbf{Provide aggregates of social media comments}: If the excerpt explicitly mentions that released artefacts would be leveraged to summarize trends or sentiments from social media comments.
    \item \textbf{Filter system outputs}: If the excerpt explicitly mentions that released artefacts would be leveraged to refine the outputs of other machine learning systems.
    \item \textbf{Maintain consistency with knowledge base}: If the excerpt explicitly mentions that released artefacts would be leveraged to ensure outputs align with established facts.
    \item \textbf{Analyze data}: If the excerpt explicitly mentions that released artefacts would be leveraged to study patterns in misinformation datasets.
    \item \textbf{Produce misinformation}: If the excerpt explicitly mentions that released artefacts would be leveraged to generate false claims for research purposes.
    \item \textbf{Vague persuasion}: If the excerpt explicitly mentions that released artefacts would be leveraged to influence user decisions without specifying the method.
\end{itemize}

If none of these labels apply, respond with "not specified in this paragraph."

\subsubsection{Ends}
\textbf{Task:} Identify the \textbf{ends} explicitly mentioned by the authors in the provided excerpt. The ends represent the ultimate goals or outcomes of the released artefacts, reflecting how the authors envision their contributions to society. The possible ends will be outlined below, and your answer should solely be based on the provided excerpt.

\paragraph{Key rules:}
\begin{enumerate}
    \item \textbf{Explicit mention only}: Consider only ends explicitly named or described in the excerpt.
    \item \textbf{Focus on the released artefacts}: Differentiate between general background information and what the authors specifically propose.
    \item \textbf{Default to 'not specified in this paragraph'}: If no ends are mentioned in the excerpt, respond with \textbf{"not specified in this paragraph"}.
    \item \textbf{Consider the big picture}: Identify how the authors envision the impact of their released automated fact-checking artefacts in the real world.
\end{enumerate}

\textbf{Question:} Based on the provided excerpt and using the provided labels, who are the explicitly mentioned ends?
\noindent
You will select the single most appropriate label that answers the question. 
Provide reasoning then choose only from the following list:

\begin{itemize}
    \item \textbf{Limit misinformation}: If the excerpt explicitly mentions that the ultimate goal of the released artefacts would be to prevent misinformation from spreading.
    \item \textbf{Limit AI-generated misinformation}: If the excerpt explicitly mentions that the ultimate goal of the released artefacts would be to prevent AI-generated falsehoods.
    \item \textbf{Increase veracity of published content}: If the excerpt explicitly mentions that the ultimate goal of the released artefacts would be to ensure accuracy of published material.
    \item \textbf{Develop knowledge of NLP/language}: If the excerpt explicitly mentions that the ultimate goal of the released artefacts would be to advance understanding of language and NLP methods.
    \item \textbf{Avoid biases of human fact-checkers}: If the excerpt explicitly mentions that the ultimate goal of the released artefacts would be to achieve unbiased and consistent fact-checking.
    \item \textbf{Detect falsehood for law enforcement}: If the excerpt explicitly mentions that the ultimate goal of the released artefacts would be to enable law enforcement to identify and act on falsehoods.
\end{itemize}

If none of these labels apply, respond with "not specified in this paragraph."
\newpage
\subsection{Research narrative classification: Hate Speech Detection}
\noindent You will be provided with the introduction of a computer science research paper on hate speech detection. Your task is to identify the research narratives that describe the intended uses of the artefacts presented in the study. You will be provided with the introduction of the study.
\vspace{0.5cm}

\noindent\textbf{Introduction:} 

\texttt{\{introduction\}}

\noindent \textbf{Task:} 
Your task is to select the research narratives that best describe the intended uses of the artefacts (e.g., datasets, models) introduced in the paper.
\begin{itemize}

    \item \textbf{Reasoning:} Provide an explanation of why each selected narrative(s) apply based on the introduction.
    \item \textbf{Selected Labels:} Choose the most appropriate label(s) from the list below. If more than one applies, separate them with commas.

Available research narratives:

\subsubsection{LLM Safety}

\textbf{Definition:} The paper discusses limiting toxic content generated by Large Language Models (LLMs). The system is designed to filter or prevent the publishing of toxic or harmful content.

\textbf{Example:}

\textit{LLMs can generate unsafe content, which harms users. We design a hate speech classifier to serve as a filter on the output of our LLM.}

\textbf{Mapping:} This narrative typically involves the end goal of limiting AI-generated toxicity or misinformation.

\subsubsection{Vague Identification}

\textbf{Definition:} The paper mentions identifying or detecting toxic posts as the means, and fighting hate as the end. However, it is unclear how the authors intend to accomplish that end using those means.

\textbf{Example:}

\textit{Hate speech is a significant issue on social media, leading to real-world harm. It is therefore crucial to develop models capable of identifying and detecting toxic content.}

\textbf{Mapping:} This narrative is identified when the application means focus on identifying toxic content, but data actors are not specified.

\subsubsection{Vague Moderation}

\textbf{Definition:} The paper proposes to assist the moderation process but does not explain how. The connection between the proposed system and how it assists moderators remains unclear.

\textbf{Example:}

\textit{Models will be used for debunking (but not how).}

\textbf{Mapping:} This narrative typically appears when the paper mentions assisting moderators but does not specify exactly how the assistance works. E.g. it is not specified whether a human will be involved in the moderation process or whether the tool is designed to automatically removing hate posts/supplant human moderators.

\subsubsection{Vague Opposition}

\textbf{Definition:} The paper presents a narrative of opposition to online hate speech without explicitly mentioning how this opposition will be enacted.

\textbf{Example:}

\textit{Hate speech is a major societal problem, eroding community trust and leading to real-world crimes. It is therefore of paramount importance that the spread of hate speech is stopped. Automated classification of hateful messages represents one solution to this critical problem.}

\textbf{Mapping:} This narrative is identified when the application means are not specified, but the ends include fighting hate.

\subsubsection{Assisted Content Moderation}

\textbf{Definition:} The paper proposes the deployment of automated moderation as a tool to assist content moderators on social media platforms. The system provides suggestions for posts to delete, but the final decision is left to human moderators.

\textbf{Example:}

\textit{Social media is rife with hate speech, eroding community trust and harming users. One solution is for moderators to remove harmful content. However, with the volume of posts made every day, this strategy is too costly. In this paper, we develop an automated system for filtering posts, helping moderators quickly discover and make decisions on circulating content.}

\textbf{Mapping:} This narrative typically includes human moderators as data actors that are involved in the moderation process (no full automation of the process).

\subsubsection{Automatic Content Moderation}

\textbf{Definition:} The paper proposes replacing human content moderators with fully automated systems. The means involve deploying classifiers to automatically detect and remove flagged posts.

\textbf{Example:}

\textit{Social media is rife with online hate, leading to real-world consequences. A costly but effective solution is to remove flagged content. In this paper, we develop an automated system for detecting harmful posts, which serves as a first line of defense against online hate.}

\textbf{Mapping:} This narrative typically includes application means focused on automated removal, supplanting human fact-checkers or generating counter narratives.

\subsubsection{Assisted Knowledge Curation}

\textbf{Definition:} The paper proposes toxic content detection primarily as a filtering component for curated knowledge vaults such as Wikipedia or knowledge graphs. The goal is to prevent the addition of harmful content.

\textbf{Example:}

\textit{Knowledge bases fuel many real-world NLP applications, such as question answering. The maintenance of knowledge bases is an expensive process, yet as knowledge evolves, they must be kept up-to-date. Automated hate speech detection systems could be used to prevent harmful content from being added or to flag existing content for review.}

\textbf{Mapping:} This narrative typically includes engineers and curators as data actors, with application means focused on filtering system outputs to maintain high-quality knowledge bases.

\subsubsection{Scientific Curiosity}

\textbf{Definition:} The paper justifies its research purely on the basis of scientific curiosity, focusing on developing better NLP models without a specific application goal.

\textbf{Example:}

\textit{Online moderation is a complex task requiring reasoning about disputed content. For systems to accurately detect harmful language, significant linguistic understanding is necessary. As such, automated moderation is an ideal testbed for developing and evaluating new NLP models.}

\textbf{Mapping:} This narrative typically includes scientists as data actors or a certain analysis, with the primary end being to develop knowledge of NLP/language.

\subsubsection{Vague Data Analysis}

\textbf{Definition:} The paper justifies its data analysis as contributing to scientific knowledge while also claiming that the goal is to fight hate, but without a clear connection between analysis and action.

\textbf{Example:}

\textit{Hate speech harms social media users. It is therefore crucial to understand the semantics of common hateful phrases. We create and analyze a dataset to this end.}

\textbf{Mapping:} This narrative typically includes application means focused on analyzing data or corpora analysis but lacks a clear connection between this analysis and real-world interventions.

\subsubsection{Truth-Telling for Law Enforcement}

\textbf{Definition:} The paper proposes hate speech detection primarily for use in law enforcement contexts, such as detecting potential threats or identifying individuals engaged in online hate speech.

\textbf{Example:}

\textit{Hate speech is illegal in some jurisdictions, but the police do not have time to monitor all instances. We develop a classifier for social media posts that can be used to identify potential perpetrators.}

\textbf{Mapping:} This narrative typically includes law enforcement as data actors, with the end goal of detecting hate speech for use in investigations.

\end{itemize}

\noindent\textbf{Prediction of the ranking model:} 

\texttt{\{Ranking explanations split by confidence levels\}}

\noindent Using your judgment and reasoning, identify the narrative(s) in the provided excerpt.
\subsection{Research narrative classification: Automated Fact-checking}

\noindent You will be provided with the introduction of a computer science research paper on automated fact-checking. Your task is to identify the research narratives that describe the intended uses of the artefacts presented in the study. You will be provided with the introduction of the study.
\vspace{0.5cm}

\noindent\textbf{Introduction:} 

\texttt{\{introduction\}}

\noindent \textbf{Task:} 
Your task is to select the research narratives that best describe the intended uses of the artefacts (e.g., datasets, models) introduced in the paper.
\begin{itemize}

    \item \textbf{Reasoning:} Provide an explanation of why each selected narrative(s) apply based on the introduction.
    \item \textbf{Selected Labels:} Choose the most appropriate label(s) from the list below. If more than one applies, separate them with commas.

Available research narratives:

\subsubsection{Vague identification}
\textbf{Definition:} The paper mentions identifying or detecting misinformative claims as the means, and limiting misinformation as the end. However, it is not clear how the authors intend to accomplish that end using those means. Typically, there are no data actors – it is also not clear who should act on the model’s predictions. Vague identification applies only in cases where the authors say that they want to identify or detect misinformation without saying what they are going to do with that afterward, e.g., in rumor detection papers where they claim that they want to detect rumors but we do not know what they want to do with this identification or classification labels.

\textbf{Example:} 

\textit{Misinformation is a major societal problem, eroding community trust and costing lives by e.g., inducing hesitance to adopt life-saving vaccines. The amount of messages spread on social media has increased drastically in recent years. Therefore, it is necessary to automatically identify potentially misinformative claims in order to address this problem.}

\textbf{Mapping:} This narrative is typically identified when the application means focus on identifying claims, but data actors are not specified.

\subsubsection{Automated External Fact-Checking}

\textbf{Definition:} When the paper proposes fully automated external fact-checking, including all parts of the pipeline. It is only automated external fact-checking when the authors explicitly say the process should be automated; otherwise, it is vague debunking.

\textbf{Example:} 

\textit{Misinformation is a major societal problem, eroding community trust and costing lives by e.g., inducing hesitance to adopt life-saving vaccines. An important way to fight misinformation is the production of relevant counter-messaging, i.e., the work done by organizations such as Full Fact or PolitiFact. With the number of false claims published on social media every hour, it is not feasible for human journalists to debunk them all. As such, the process must be automated.}

\textbf{Mapping:} This narrative typically includes application means designed to supplant human fact-checkers, automating the entire fact-checking process.

\subsubsection{Vague Opposition}

\textbf{Definition:} Restricted to cases without any application means (model means and no application means in contrast to vague identification and vague debunking). E.g., a machine learning/automated system will reduce the spread of misinformation. When the paper presents a narrative of vague opposition to misinformation. The end is to limit the spread or influence of misinformation. However, the connection between modeling means and ends is left unmentioned. An impression is given that the development of automated fact-checking will limit the spread of misinformation, but the link between the two is left unstated.

\textbf{Example:} 

\textit{Misinformation is a major societal problem, eroding community trust and costing lives by e.g., inducing hesitance to adopt life-saving vaccines. It is therefore of paramount importance that the spread of false information is stopped. Automated fact-checking – that is, the automatic classification of claim veracity – represents one solution to this critical problem.}

\textbf{Mapping:} Application means are typically not specified, but the ends typically include limiting misinformation or limiting AI-generated misinformation.

\subsubsection{Assisted External Fact-Checking}

\textbf{Definition:} When the paper proposes to deploy automated fact-checking as an assistive tool for journalists, deployed for external fact-checking. This is restricted to when the paper proposes to improve one or multiple components in the automated fact-checking pipeline rather than the whole pipeline/end-to-end system (in the latter case it becomes automated external fact-checking). I.e., when it is human-in-the-loop, then it is assisted fact-checking but not automated.

\textbf{Example:} 

\textit{Misinformation is a major societal problem, eroding community trust and costing lives by e.g., inducing hesitance to adopt life-saving vaccines. An important way to fight misinformation is the production of relevant counter-messaging, i.e., the work done by organizations such as Full Fact or PolitiFact. With the number of false claims published on social media every hour, it is not feasible for human journalists to debunk them all. Journalists could use automated fact-checking to triage incoming claims to limit the workload, or to quickly surface relevant evidence while producing articles.}

\textbf{Mapping:} This narrative typically involves data actors such as professional journalists, citizen journalists, or technical writers, supported by modeling means like human-in-the-loop, classify/score veracity, or evidence retrieval.

\subsubsection{Vague Debunking}

\textbf{Definition:} The authors propose to produce evidence-based debunkings of text via automated means, but it is not clear whether the introduced model is an assistive tool for fact-checkers or fully automated. Further, it is not clear how the “debunkings” will be communicated to misinformation-believers. It is clear that the mechanism is supposed to follow what fact-checkers are currently doing, but not where or how the ML model will be used in this process. Furthermore, it is unclear whether the entire process will be automated. For example, the paper may suggest that an automated fact-checking model should be used by fact-checkers, but it is not clear how the model assists in that.

\textbf{Example:}

\textit{Misinformation is a major societal problem, eroding community trust and costing lives by e.g., inducing hesitance to adopt life-saving vaccines. One proposed solution is to debunk circulating claims, i.e. to find evidence against them. This is the process commonly carried out by fact-checking organizations, e.g. PolitiFact. In this paper, we introduce a classifier...}

\textbf{Mapping:} This narrative may typically include application means like gathering and presenting evidence or providing labels/veracity scores, supported by modeling means like classify/score veracity or evidence retrieval. However, data actors are not specified, and it remains unclear how the debunking outputs will be used.

\subsubsection{Assisted Media Consumption}

\textbf{Definition:} If the paper proposes to deploy automated fact-checking as an assistive tool for consumers of information, either as a layer adding extra information to social media posts or as a standalone site where claims can be tested.

\textbf{Example:}

\textit{Misinformation is a major societal problem, eroding community trust and costing lives by e.g. inducing hesitance to adopt life-saving vaccines. It is therefore of paramount importance that the spread of false information is stopped. Studies have shown that many people adopt beliefs without doing due diligence on the information they receive. Automated fact-checking – that is, the automatic classification of claim veracity – could via e.g., a plugin be deployed to warn social media users about potentially false claims.}

\textbf{Mapping:} This narrative may typically feature model owners as social media companies, and data subjects as social media users. The application means can include vague persuasion techniques or aim to limit misinformation by subtly influencing user interaction with content. Alternatively, data actors may be media consumers directly benefiting from these systems.

\subsubsection{Scientific Curiosity}

\textbf{Definition:} When the authors of the paper justify their projects purely based on scientific curiosity. While differing strongly from the other narratives presented here, this is still a virtue epistemic narrative, concerned with the production of good knowledge.

\textbf{Example:}

\textit{Journalistic fact-checking is a difficult task, requiring reasoning about disputed claims that fool sufficiently many humans to warrant professional attention. For systems to mimic fact-checking to a substantial degree, significant semantic understanding is necessary. As such, automated fact-checking is an ideally suited field to develop and test new models for natural language understanding.}

\textbf{Mapping:} This narrative may typically involve model owners and data actors who are scientists. The end is to develop knowledge of NLP/language, focusing on advancing scientific understanding rather than practical applications.

\subsubsection{Assisted Knowledge Curation}

\textbf{Definition:} When the paper proposes fact-checking primarily as a component filtering the information kept in some curated knowledge vault, including graph-based knowledge bases as well as text-based collections such as Wikipedia.

\textbf{Example:}

\textit{Knowledge bases fuel many real-world NLP applications, e.g., question answering. The maintenance of knowledge bases is an expensive process, yet as new facts appear in the world knowledge bases must be kept up-to-date. Automated triple extraction from e.g., news data has been proposed as an alternative to human annotators; yet, the quality remains low. Automated fact-checking systems, which verify facts against trusted knowledge sources, could be used to prevent highly disputed facts from being entered – or ensure that new facts are consistent with the existing knowledge.}

\textbf{Mapping:} This narrative may typically include data actors like engineers and curators who use application means to filter system outputs and modeling means such as classify/score veracity, evidence retrieval, or human-in-the-loop techniques. The end is to increase the veracity of published content, ensuring high-quality information in curated knowledge bases.
\subsubsection{Adversarial Research}

\textbf{Definition:} When the paper proposes the development or use of automated fact-checking (AFC) systems to explore vulnerabilities in misinformation detection, either by generating adversarial examples, repurposing real content in misleading ways, or testing system robustness. The focus is on exposing weaknesses in fact-checking frameworks or demonstrating new, sophisticated threat scenarios (e.g., mismatched image-caption pairs, AI-generated deepfakes).

\textbf{Mapping:} Adversarial research typically occurs when data actors are scientists, who employ modeling means to generate claims and use application means to produce misinformation.
\subsubsection{Assisted Internal Fact-Checking}

\textbf{Definition:} When the paper proposes to deploy automated fact-checking as an assistive tool for journalists, deployed internally.

\textbf{Example:}

\textit{Research is a fundamental task in journalism, conducted to ensure published information is truthful and to protect the publisher from libel suits. This is a crucial step, which journalists – strained by the advent of the 24-hour news cycle – increasingly skip. Given a trusted source of evidence documents, such as LexisNexis, much of the grunt work of research could be handled by automated fact-checkers, leaving journalists free to tackle the hardest parts, e.g. double-checking information with sources.}

\textbf{Mapping:} This narrative may typically feature data actors and data subjects as professional journalists, citizen journalists, or technical writers, using application means that do not supplant human fact-checkers but help to increase the veracity of published content or limit misinformation. This narrative emphasizes internal verification processes within journalistic organizations before content publication.

\subsubsection{Automated Content Moderation}

\textbf{Definition:} If the paper analyzed proposes a similar content moderation strategy, but instead of assisting human moderators, it suggests replacing them entirely. In this case, the end is to limit the spread of misinformation on social media platforms, the means is to deploy classifiers to truth-tell claims and remove any labeled false, and the actors are the algorithm (as well as, implicitly, the model owners – the executives and engineers at social media companies who deploy and make decisions about such systems).

\textbf{Example:}

\textit{Social media is rife with misinformation, eroding community trust and costing lives by e.g. inducing hesitance to adopt life-saving vaccines. One solution is for moderators to remove information deemed false. However, with the number of posts made every day on social networks, this strategy is too costly. In this paper, we develop an automated system for detecting false claims, which can serve as a first line of defense against misinformation.}

\textbf{Mapping:} This narrative may typically feature data actors as algorithms, model owners as social media companies, and data subjects as social media users. The application means can also involve automated removal of content to reduce misinformation without human intervention.

\subsubsection{Truth-Telling for Law Enforcement}

\textbf{Definition:} When the paper proposes fact-checking primarily as a lie detector for use in law enforcement contexts. This could include police work as well as courtroom applications.

\textbf{Example:}

\textit{Automatic detection of deception is commonly used in police work via e.g. polygraphs. However, accuracy remains low. We propose that automated fact-checking via NLP could represent a viable alternative.}

\textbf{Mapping:} In this narrative, data actors are typically law enforcement, utilizing systems with the end to detect falsehood for law enforcement. These tools support investigations, legal processes, or forensic analysis.

\end{itemize}
\noindent\textbf{Prediction of the ranking model:} 

\texttt{\{Ranking explanations split by confidence levels\}}

\noindent Using your judgment and reasoning, identify the narrative(s) in the provided excerpt.
\end{document}